\PassOptionsToPackage{table}{xcolor}
\documentclass{bmvc2k}
\usepackage{subcaption}
\usepackage[table]{xcolor}

\usepackage{subcaption}
\usepackage{relsize}
\usepackage{sidecap}
\usepackage{caption} 
\usepackage{amssymb}
\usepackage{graphicx}
\usepackage{colortbl}
\usepackage{arydshln}
\usepackage{multirow}
\usepackage{booktabs}
\usepackage{algorithm}
\usepackage[noend]{algpseudocode}
\usepackage{wrapfig}

\definecolor{lightblue}{RGB}{173,216,230}  
\definecolor{lightgreen}{RGB}{240, 255, 240}
\definecolor{verylightgray}{RGB}{240, 240, 240}
\definecolor{darkgreen}{rgb}{0.0, 0.4, 0.0}
\definecolor{darkred}{rgb}{0.75, 0.0, 0.0}
\definecolor{green_recursive}{HTML}{D7F5E1}
\definecolor{purple_recursive}{HTML}{F6E2F4}
\definecolor{blue_recursive}{HTML}{DCEAF7}
\definecolor{yellow_recursive}{HTML}{FFFFD5}

\title{ETTA: Efficient Test-Time Adaptation for Vision-Language Models through Dynamic Embedding Updates}

\addauthor{Hamidreza Dastmalchi}{hrd@yorku.ca}{1}
\addauthor{Aijun An}{aan@yorku.ca}{1}
\addauthor{Ali Cheraghian}{ali.cheraghian@data61.csiro.au}{2,3}

\addinstitution{
 York University\\
 Toronto, Canada
}
\addinstitution{
 The Australian National University\\
  Canberra, Australia
}

\addinstitution{
  Data61-CSIRO, Australia
}

\runninghead{Dastmalchi et al}{ETTA}

\begin{document}

\maketitle

\begin{abstract}
Pretrained vision-language models (VLMs) like CLIP show strong zero-shot performance but struggle with generalization under distribution shifts. Test-Time Adaptation (TTA) addresses this by adapting VLMs to unlabeled test data in new domains. While some TTA methods rely on prompt-tuning, training-free cache-based approaches are preferred for efficiency. However, current cache-based TTA models store only a limited set of high-confidence samples, restricting the decision boundary to these samples and ignoring the influence of other incoming test data. To address this, we propose Efficient Test-Time Adaptation (ETTA), introducing a Recursive Updating module that integrates all incoming test samples, progressively refining the decision boundary. This strategy mimics an unbounded cache, dynamically updating contextual embeddings for improved accuracy with minimal memory and computational overhead. ETTA also includes an Adaptive Ensemble module to reduce prompt dependency in image-to-text scores by dynamically selecting optimal prompts for each class. Furthermore, ETTA adaptively combines scores from both modules based on confidence levels, leveraging their complementary strengths. Extensive experiments on two benchmarks confirm that ETTA surpasses the state-of-the-art TTA models in computational complexity and accuracy, setting a new standard for effective, efficient test-time adaptation. The code has been released at \url{https://github.com/hamidreza-dastmalchi/ETTA}.
\end{abstract}

\section{Introduction}
\label{sec:intro}

Vision-Language Models (VLMs) like CLIP \cite{clip} and ALIGN \cite{align} excel at merging visual and textual information for tasks such as classification \cite{gondal2024domain, Huang_2023_ICCV, jeong2023winclip}, generation \cite{styleclip, clipdraw, glide, diff-hierarchical-clip}, and recognition \cite{clip-action-recog, actionclip, pointclip, chen2024towards}. CLIP’s core strength lies in its ability to generalize to new tasks without fine-tuning, enabled by training on large-scale image-text pairs that align visual and textual features in a shared embedding space via contrastive learning. Using crafted prompts like “a photo of a \textless category\textgreater,” CLIP generates text embeddings that match visual embeddings, enabling classification across tasks it was not explicitly trained for.

\begin{SCfigure}
\vspace{.1cm}
\hspace{-.4cm}
\includegraphics[width=0.5\linewidth,trim=0cm 0cm 0cm 0cm, clip]{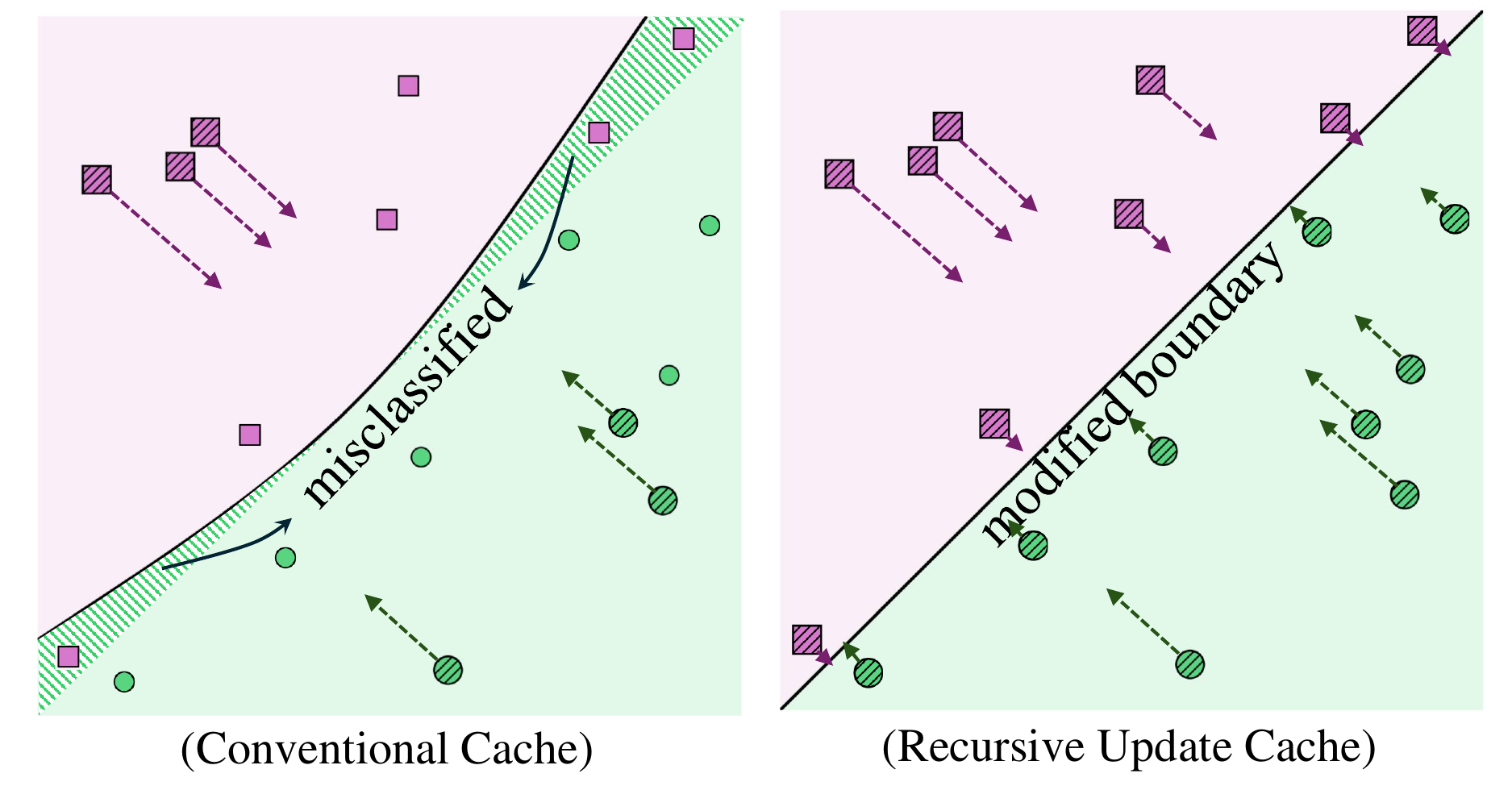}
\caption{\small \textbf{(Left):} Conventional cache-based adaptation, where a few high-confidence prototypes (shown as larger data points) define the decision boundary, often leading to misclassification of samples near the boundary. \textbf{(Right):} Our Recursive Update cache incorporates all data points in forming the decision boundary, resulting in enhanced classification performance.}
\label{fig:etta-motivation}
\label{fig:etta-motivation}
\end{SCfigure}

Inspired by prompt learning in NLP \cite{li2021prefix, lester2021power, gao2020making}, researchers have introduced prompt tuning as an efficient way to align CLIP with various downstream tasks. However, prompt tuning requires labeled data, which is often unavailable in real-world scenarios. To mitigate this limitation, Test-time Prompt Tuning (TPT) ~\cite{tpt} was introduced to enhance zero-shot classification by minimizing prediction entropy across image augmentations. C-TPT \cite{C-TPT} builds upon TPT, addressing its calibration issue. However, the computational demand of optimizing learnable prompts through backpropagation remains misaligned with the efficiency requirements of test-time tasks. To address this issue, training-free adaptation strategies~\cite{TDA, DMN, Sus-x, CuPL} have been proposed, enabling VLMs to adapt to new samples at test time without backpropagation. Nevertheless, the training-free TTA approaches in \cite{Sus-x} and \cite{CuPL} rely on external models, specifically diffusion models \cite{ddpm} and large language models (LLMs), respectively.  In contrast, cache-based methods \cite{TDA, DMN} are more efficient and independent of external models. These methods store a limited number of highly confident test samples, allowing for the computation of affinities within the image embedding space.

Cache-based TTA approaches are efficient but rely exclusively on a small set of highly confident test samples for predictions, limiting the model’s adaptability. In Fig. \ref{fig:etta-motivation} (Left), this limitation is demonstrated using a 2D toy example, where only three prototypes per class (shown as larger data points) define the decision boundary. As a result, this boundary depends solely on these prototypes, leading to misclassification within the hashed region (indicated by a shaded area in the figure). In contrast, we propose a Recursive Updating strategy in which all points contribute to forming the decision boundary, as illustrated in Fig. \ref{fig:etta-motivation} (Right), resulting in more reliable classifications. Our approach achieves accuracy equal to an unbounded cache while requiring significantly less memory and computational overhead.

Additionally, cache-based TTA models have another limitation: they still rely on prompts. While combining cache scores with VLM scores improves predictions, the latter are highly sensitive to prompt wording and vary across domains and classes. Although prompt ensembling \cite{clip} aims to improve robustness, our experiments show it does not consistently enhance performance. To address this, we propose Adaptive Ensembling, which filters out irrelevant prompts based on their similarity to the input image.

To this end, we introduce Efficient Test-Time Adaptation (ETTA) to address the prompt dependency and limited adaptability of the conventional cache-based model. ETTA comprises three key components: the \textbf{Adaptive Ensemble} module, the \textbf{Recursive Update} module, and an \textbf{Adaptive Fusion} module. The Adaptive Ensemble module enhances the reliability of image-to-text scores by reducing sensitivity to prompt quality. Similar to \cite{clip}, we prepend a set of templates with class names to create class-specific prompts. However, rather than simply ensembling the scores from these prompts, our module filters out prompts less representative of each class based on their similarity to the input image. It then averages the embeddings of only the most relevant prompts, leading to more accurate predictions for each test sample. Additionally, the Recursive Update module matches the performance of an unbounded cache by progressively refining the decision boundary using all incoming test samples. Unlike traditional caches that store multiple prototypes, it maintains just one contextual embedding per class, updated recursively. This design minimizes memory and computation while preserving high accuracy. Finally, the Adaptive Fusion module adaptively combines the scores from these two modules, weighting them based on their confidence.

In summary, the contributions of this work are as follows:
\textbf{(1)} We introduce an Adaptive Ensemble module that enhances image-to-text score adaptability by selecting and averaging the most relevant prompts for each incoming image. \textbf{(2)} We propose a Recursive Update module that iteratively refines contextual embeddings by dynamically incorporating all arriving test samples, achieving accuracy comparable to an unbounded cache. This approach drastically reduces storage and computation while surpassing the accuracy of conventional cache-based models. \textbf{(3)} We extensively evaluate our method on two benchmarks, demonstrating superior accuracy and efficiency over state-of-the-art approaches.


\section{Related Work}
\label{sec:related-work}
\noindent\textbf{Vision-Language Models Adaptation:} Vision-language models \cite{clip, align, li2019visualbert, alayrac2022flamingo, chen2020uniter, TCL, li2022blip, li2023blip} are highly effective at capturing complex semantics from large-scale image-text datasets. CLIP \cite{clip}, for instance, aligns visual and textual representations in a shared embedding space, achieving impressive zero-shot performance without task-specific fine-tuning. Inspired by NLP \cite{lester2021power, li2021prefix, P-tuning}, prompt tuning further improves task-specific adaptation by learning prompts in a continuous embedding space. Prompt tuning can be applied to the language side (e.g., CoOp \cite{coop}, CoCoOp \cite{cocoop}), the vision side \cite{visual-prompt, wang2022dualprompt-visual, wang2022learning-visual-prompt}, or both \cite{khattak2023maple, prompt-src}, enhancing model flexibility. While prompt tuning and other adaptation techniques like CLIP-Adapter \cite{Clip-adapter} and TaskRes \cite{Task-residual} improve performance, they rely on annotated downstream data, limiting their practicality in real-world settings. In contrast, this work introduces test-time adaptation, enabling models to adjust dynamically at inference without training data.

\vspace{0.2cm}
\noindent\textbf{Test Time Adaptation:} Test-time adaptation (TTA) enables models to adjust to new, unseen data without additional labeled data or retraining. Numerous methods have been proposed to address test-time adaptation challenges. Common techniques include adapting batch normalization statistics \cite{schneider2020improving}, updating selective model weights \cite{partial-weights1, partial-weights2, partial-weights3}, or combining both \cite{zhang2021adaptive, Tent} to better align with test data distributions. Other strategies involve minimizing prediction entropy to encourage confident outputs \cite{Tent, confidence-maximization}, generating source-like data to support accurate predictions \cite{gao2023back, tsai2024gda, song2023target, dastmalchi2025test}, and leveraging self-supervised training to improve generalization \cite{masked-autoencoder, self-supervised1}. Augmentations have been used to enforce prediction consistency at test time, improving model robustness \cite{zhang2022memo, do-we-need-prompt}. Yet, test-time adaptation for vision-language models remains underexplored. Test-time Prompt Tuning (TPT) \cite{tpt} optimizes prompts to reduce prediction entropy across augmentations. DiffTPT \cite{difftpt} extends this using a pretrained diffusion model for richer augmentations, while SwapPrompt \cite{swapprompt} employs contrastive learning for prompt adaptation. PromptAlign \cite{promptalign} aligns token-level statistics to bridge the distribution gap between test and source samples.

\vspace{0.2cm}
\noindent\textbf{Training-Free Test-time Adaptation:} SuS-X \cite{Sus-x} and CuPL \cite{CuPL} are training-free approaches that leverage external models for test-time adaptation of vision-language models. In contrast, cache-based models use prototypes for non-parametric test-time predictions, storing high-confidence samples to compute affinities with new test data \cite{TDA, DMN, adanpc, T3A, tip-adapter, grave2017unbounded}. Unbounded cache methods, like \cite{grave2017unbounded}, manage long-term dependencies by selectively retaining key contexts but still face memory management challenges. Tip-Adapter \cite{tip-adapter} adapts CLIP in few-shot settings using cached prototypes but requires labeled data. Unlike Tip-Adapter, TDA \cite{TDA} and DMN \cite{DMN} support dynamic, training-free adaptation to test-time distributions without labeled data. However, these methods rely on a limited set of prototypes, which restricts the influence of other test samples on refining the decision boundary, potentially limiting adaptability to new data variations.

\section{Method}
\subsection{Preliminaries}
\textbf{Contrastive Language-Image Pre-training (CLIP):} CLIP \cite{clip} comprises a text encoder \(\operatorname{E_t}\) and a vision encoder \(\operatorname{E_v}\) for different modalities. For zero-shot classification, an image \(\mathbf{I} \in \mathbb{R}^{c \times h \times w}\) with class label \( y \in \{1, \ldots, C\} \) is processed by the vision encoder, producing an embedding \( \mathbf{v} = \operatorname{E_v}(\textbf{I}) \in \mathbb{R}^d \). A template, such as \( t = \text{``a photo of a "} \), is prepended to each class name, forming prompts \( p_i\), which are encoded by the text encoder to produce class embeddings \( \mathbf{w}_i = \operatorname{E_t}(p_i) \in \mathbb{R}^d \). Finally, the similarity between the image embedding and each class-specific text embedding is computed to assign class probabilities.

\label{sec:method}
\begin{equation}
    p\left(y_i \mid {\mathbf{I}}\right)=\frac{\exp \left(\text{sim}(\mathbf{w}_{{i}} , \mathbf{v}) /\tau\right)}{\sum_{i=1}^C \exp \left(\text{sim}( \mathbf{w}_{{i}} , \mathbf{v}) /\tau\right)}
    \label{eq:probability}
    \vspace{-0.1cm}
\end{equation}

\noindent where $\text{sim}(\cdot)$ represents cosine similarity, and $\tau$ is the temperature parameter.

\subsection{Problem Formulation}

At test time, we receive unlabeled images in a stream, denoted as \(\mathbf{I} \in \mathcal{Q}_t\), where \(\mathcal{Q}_t\) is the target domain with a distribution shift from the original training domain \(\mathcal{Q}_s\). Labeled image-text pairs from the training domain are no longer accessible. The goal is to adapt model predictions using only unlabeled samples from \(\mathcal{Q}_t\) to handle this shift.

\begin{figure*}[!t]
    \centering
    \includegraphics[width=0.86\linewidth]{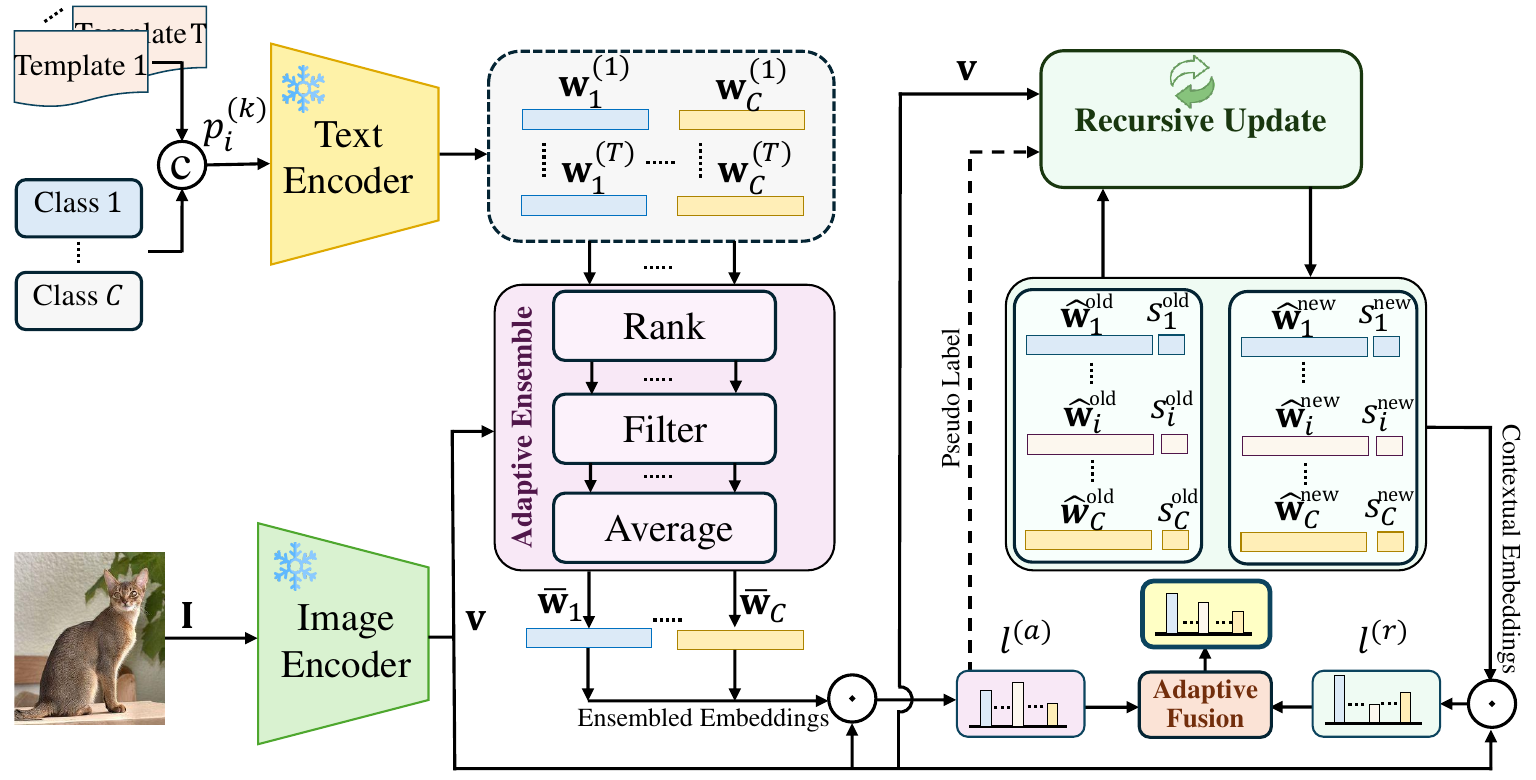}
    \vspace{0.25cm}
    \caption{\small The image \(\mathbf{I}\) is encoded into a vision embedding \(\mathbf{v}\), while \(C\) class names with \(T\) prompts are encoded into text embeddings \(\{ \mathbf{w}_i^{(k)} \}_{i=1}^C{}_{k=1}^T\). The Adaptive Ensemble filters and averages relevant text embeddings to form \(\{ \overline{\mathbf{w}}_i \}_{i=1}^C\), yielding adaptive score \(l^{(a)}\). The Recursive Update refines contextual embeddings \(\{ \widehat{\mathbf{w}}_i \}_{i=1}^C\), producing recursive score \(l^{(r)}\). The final prediction fuses \(l^{(a)}\) and \(l^{(r)}\).
}
    \label{fig:clip-tta-blockdiagram}
    \vspace{-0.4cm}
\end{figure*}


\subsection{ETTA: Efficient Test-Time Adaptation}

Our model consists of three modules, shown in Fig.~\ref{fig:clip-tta-blockdiagram}: (a) \textbf{Adaptive Ensemble}, which refines class-specific prompts based on the input image; (b) \textbf{Recursive Update}, which recursively updates contextual prompts from the data stream; and (c) \textbf{Adaptive Fusion}, which combines logits from both modules to produce the final prediction.

\vspace{0.2cm}
 \noindent{\textbf{Adaptive Ensemble:}} To improve robustness, the Adaptive Ensemble module dynamically selects relevant prompts for each test sample. Following CLIP \cite{clip}, we use a standardized set of generic templates, such as “a photo of many,” across all classes. Given \( T \) templates, we generate \( T \) class-specific prompts by appending each class name to each template \( t^{(k)} \), forming \( p_i^{(k)} = \{t^{(k)}; \text{name}_i\} \), as illustrated in Fig. \ref{fig:clip-tta-blockdiagram}. Each \( p_i^{(k)} \) is then processed by the frozen text encoder \( \operatorname{E_t} \) and normalized to produce the class-specific representation.


\begin{equation}
  \mathbf{w}_i^{(k)} = \frac{\operatorname{E_t}(p_i^{(k)})}{\|\operatorname{E_t}(p_i^{(k)})\|}, \quad k = 1, \ldots, T, \quad i = 1, \ldots, C
\end{equation}

\vspace{-0.2cm}
\noindent where \(\mathbf{w}_i^{(k)} \in \mathbb{R}^{ d}\) represents the embedding for the \(k\)-th prompt of class \(y_i\). Similarly, the image arriving on the fly (\(\mathbf{I}\)) is passed through the frozen vision encoder \( \operatorname{E_v} \) and subsequently normalized to construct the image representation \( \mathbf{v} \in \mathbb{R}^{d} \): \(  \mathbf{v} = \frac{\operatorname{E_v}(\mathbf{I})}{\|\operatorname{E_v}(\mathbf{I})\|}    
\)


Next, the \( T \) text embeddings for each class are ranked based on their cosine similarity to the input image embedding. We retain the top \(\alpha\) percentile and discard the rest as outliers.

\begin{equation}
    \mathcal{W}_i^{\alpha} = \left\{ \mathbf{w}_i^{(k)} \mid \langle \mathbf{w}_i^{(k)}, \mathbf{v}\rangle \geq \gamma_{\alpha} \right\}
    \label{eq:filter}
\end{equation}

\noindent where \( \mathcal{W}_i^{\alpha} \) is the set of retained embeddings for class \( i \), and \( \gamma_{\alpha} \) represents the similarity threshold for the top \(\alpha\) percentile. The ensembled class embedding is then computed by averaging and normalizing the retained text embeddings:


\begin{equation}
    \overline{\mathbf{w}}_i =
    \frac{\sum\limits_{\scalebox{0.6}{$\mathbf{w}_i^{(k)} \in \mathcal{W}_i^{\alpha}$}} \mathbf{w}_i^{(k)}}
    {\bigl\| \sum\limits_{\scalebox{0.6}{$\mathbf{w}_i^{(k)} \in \mathcal{W}_i^{\alpha}$}} \mathbf{w}_i^{(k)} \bigr\|}  \quad \text{for } \, i = 1, \ldots, C
\end{equation}

This refinement process is illustrated in Fig. \ref{fig:clip-tta-blockdiagram}. The ensembled text embedding is then used to calculate the adaptive logit, \( l^{\text{(a)}}_i \):

\begin{equation}
    l^{\text{(a)}}_i  =   \langle \overline{\mathbf{{w}}}_i,\mathbf{v}\rangle,  \quad \text{for } \, i = 1, \ldots, C
\end{equation}

\noindent \textbf{Recursive Update:} To describe our Recursive Update cache, we start with a bounded cache that stores the most confident prototypes. Suppose each class \( i \) maintains a cache \( \mathbf{V} = [\mathbf{v}_1; \mathbf{v}_2; \dots; \mathbf{v}_n] \in \mathbb{R}^{n \times d} \) with the top \( n \) confident image embeddings. We contextualize class embeddings via cross-attention between the ensembled embedding \( \overline{\mathbf{w}}_i \) and its corresponding prototypes:

\begin{equation}
        \widehat{\mathbf{w}}_{i} = \sigma\big(\overline{\mathbf{w}}_i \mathbf{V}^T\big)\,\mathbf{V}, \;\; \text{for } \, i = 1, \ldots, C
        \label{eq:(4)}
\end{equation}
where \(\widehat{\mathbf{w}}_{i}\) is the contextual embedding and \(\sigma\) is the Softmax function. Cross-attention computes contextual embeddings by a weighted sum of cache prototypes, with weights based on their similarity to each text embedding. In Section~\ref{sec:ablation}, we show that a larger cache improves performance but increases computation and memory costs. To address this, we propose a Recursive Updating strategy for an unbounded cache, avoiding prototype storage and cross-attention at each step. We begin by expressing Eq.~\ref{eq:(4)} in vector form:

\vspace{-0.1cm}

\begin{equation}
    \widehat{\mathbf{w}}_{i} = \frac{\sum_{k=1}^{n} \mathbf{v}_k \exp\left( \langle\overline{\mathbf{w}}_{i},\mathbf{v}_k \rangle\right)}{\sum_{k=1}^{n}\exp\left(\langle \overline{\mathbf{w}}_{i}, \mathbf{v}_k \rangle\right)}
\end{equation}

Let \( \widehat{\mathbf{w}}_i^{\text{old}} \) denote the contextual embedding before the new image arrives. After arrival, the image embedding with pseudo-label \( i \) is added to the corresponding cache, updating the contextual embedding to \( \widehat{\mathbf{w}}_i^{\text{new}} \) as follows:


\begin{equation}
    \widehat{\mathbf{w}}_{i}^{\text{new}} = \frac{s_i^{\text{old}} \,\widehat{\mathbf{w}}_{i}^{\text{old}} + \exp\left(\langle\overline{\mathbf{w}}_{i}, \mathbf{v}\rangle\right) \, \mathbf{v}}{s_i^{old} + \exp\left(\langle\overline{\mathbf{w}}_{j}, \mathbf{v} \rangle\right)}
    \label{eq:recursive1}
\end{equation}
where \( s_i^{\text{old}} \) denotes the accumulated sum of exponential terms before the new image arrives: \( s_i^{\text{old}} = \sum_{k=1}^{n} \exp\left(\langle\overline{\mathbf{w}}_{i}, \mathbf{v}_k \rangle\right) \). Subsequently, \( s_i^{n} \) is updated by incorporating the new exponential term from the arriving image embedding:
\begin{equation}
    s_i^{\text{new}} = s_i^{\text{old}} + \exp\left(\langle\overline{\mathbf{w}}_{i}, \mathbf{v}\rangle \right)
    \label{eq:recursive2}
\end{equation}

Note that the input image embedding \( \mathbf{v}\), with pseudo label \( i \), affects only the \( i \)-th class embedding; thus, \( \widehat{\mathbf{w}}_k \) and \( s_k \) for all \( k \neq i \) remain unchanged:

\begin{equation}
    \widehat{\mathbf{w}}_{k}^{\text{new}} = \widehat{\mathbf{w}}_{k}^{\text{old}}, \quad \text{and} \quad
    s_k^{\text{new}}=s_k^{\text{old}}  \quad \forall k \neq i
    \label{eq:recursive3}
\end{equation}

In other words, instead of storing samples in an unbounded cache and performing costly cross-attention, we use the efficient recursive Eqs. \ref{eq:recursive1}, \ref{eq:recursive2}, and \ref{eq:recursive3} to update only the relevant contextual text embedding at each time step based on the pseudo label of the arriving image. Finally, the set of updated contextual embeddings \( \{\widehat{\mathbf{w}}_i^{\text{new}}\}_{i=1}^{C} \) is used to compute the corresponding scores, denoted as \( l_i^{(r)} \), as depicted in Figure \ref{fig:clip-tta-blockdiagram}.
\begin{equation}
    l_i^{(\text{r})} = \frac{\langle\widehat{\mathbf{{w}}}^{\text{new}}_i,\mathbf{v}\rangle  }{\|\widehat{\mathbf{{w}}}_i^{\text{new}}\|} \quad \text{for } i = 1, \ldots, C
\vspace{-0.2cm}
\end{equation}

\vspace{0.2cm}
\noindent \textbf{Adaptive Fusion:} In the final step, the scores from the adaptive component and the recursive component are combined in inverse proportion to their respective entropies, assigning greater weight to the component with more confident predictions:
\vspace{-0.2cm}
\begin{equation}
    l^{(c)}_{i} = \frac{{H}^{(r)}}{{H}^{(a)}+{H}^{(r)}} l^{(a)}_i+\frac{{H}^{(a)}}{{H}^{(a)}+{H}^{(r)}}l^{(r)}_i , \text{for } i = 1, \ldots, C
\end{equation}
where \( H^{(a)} \) and \( H^{(r)} \) represent the entropies of the adaptive and recursive logits, respectively. The combined logits are then used to make the final class prediction.

\vspace{-0.2cm}

\begin{equation}
    \hat{y} = \underset{i}{\arg \max} \left(l_i^{(c)}\right)
\end{equation}
\vspace{-0.3cm}
The complete procedure is outlined in appendix \ref{sec:algorithm}.

\vspace{-0.1cm}

\section{Experiments}
\label{sec:experiments}


\subsection{Datasets}

Our main experiments focus on two widely used benchmarks for evaluating test-time adaptation in vision-language models \cite{tpt, difftpt, TDA}. The OOD benchmark tests generalization to unfamiliar distributions using four ImageNet variants: ImageNet-A \cite{imagenet-a}, V2 \cite{imagenet-v2}, R \cite{imagenet-r}, and S \cite{imagenet-s}. The cross-domain benchmark evaluates adaptability on test sets from ten diverse datasets across distinct domains, including Aircraft \cite{aircraft}, Caltech101 \cite{caltech}, Cars \cite{cars}, DTD \cite{dtd}, EuroSAT \cite{eurosat}, Flower102 \cite{flower}, Food101 \cite{food}, Pets \cite{pets}, SUN397 \cite{sun397}, and UCF101 \cite{ucf101}.

\subsection{Baselines}

\noindent \textbf{Source Model (SRC):} This represents the standard, unadapted CLIP model. We evaluate it using a {single generic prompt} (``a photo of a") as well as an {ensemble of 80 handcrafted generic prompts} from Radford et al. \cite{clip} denoted as CLIP and CLIP\(^*\), respectively.

\vspace{0.1cm}
\noindent \textbf{Few-Shot Adaptation (FSA):} This category includes {CoOp} \cite{coop}, {CoCoOp} \cite{cocoop}, and {Tip-Adapter} \cite{tip-adapter}, which are trained on {ImageNet} \cite{imagenet} and evaluated on both {out-of-distribution (OOD)} and {cross-domain benchmarks}. However, Tip-Adapter is only applicable to the OOD setting, as it does not generalize to unseen test-time classes.

\vspace{0.1cm}
\noindent \textbf{Generic Prompt-based TTA (GP):} This category includes methods that use generic  phrases combined with class names to form class-specific prompts (e.g., \textit{``a high-resolution photo of a [CLS]"}, where \textit{[CLS]} is the class). TDA \cite{TDA} directly adapts using generic prompts, while TPT \cite{tpt} and DiffTPT \cite{difftpt} apply prompt tuning initialized from them.

\vspace{0.1cm}
\noindent \textbf{Class-Specific Prompt-based TTA (CSP):} These methods use GPT-generated class descriptions (e.g., “The Abyssinian is a medium-sized, short-haired cat” \cite{CuPL, Sus-x, DMN}). For fair comparison, we extend our method with class-specific GPT prompts, denoted as ETTA$^+$, and evaluate it against DMN-ZS \cite{DMN}, CuPL \cite{CuPL}, and SuS-X \cite{Sus-x}.

\begin{table}[!t]
\centering
\renewcommand{\arraystretch}{1}
\setlength{\tabcolsep}{4pt}
\resizebox{0.84\linewidth}{!}{
\begin{tabular}{clccccccc}
\hline
\rowcolor{verylightgray}
& \textbf{Method} & \textbf{ImageNet} & \textbf{ImageNet-A} & \textbf{ImageNet-V2} & \textbf{ImageNet-R} & \textbf{ImageNet-S} & \textbf{Average} & \textbf{OOD Avg} \\
\hline
&  &  &  & ResNet-50 Backbone &  &  & &  \\
\hline
\multirow{2}{*}{\footnotesize \rotatebox{90}{\textbf{SRC}}} 
& \(\text{CLIP}\)-ResNet-50 \cite{clip} & 58.16 & 21.83 & 51.41 & 56.15 & 33.37 & 44.18 & 40.69 \\
& \(\text{CLIP}^*\)-ResNet-50 \cite{clip} & 59.81 & 23.24 & 52.91 & 60.72 & 35.48 & 46.43 & 43.09 \\
\hdashline
\multirow{3}{*}{\footnotesize\rotatebox{90}{\textbf{FSA}}} 
& CoOp \cite{coop} & {63.33}  & 23.06 & 55.40 & 56.60 & 34.67 & 46.61 & 42.43 \\
& CoCoOp \cite{cocoop} & 62.81 & 23.32 & 55.72 & 57.74 & 34.48 & 46.81 & 42.82 \\
& Tip-Adapter \cite{tip-adapter} & 62.03 & 23.13 & 53.97 & 60.35 & 35.74 & 47.04 & 43.30 \\
\hdashline
\multirow{4}{*}{\footnotesize\rotatebox{90}{\textbf{GP}}} 
& TPT \cite{tpt} & 60.74 & 26.67 & 54.70 & 59.11 & 35.09 & 47.26 & 43.89 \\
& DiffTPT \cite{difftpt} & 60.80 & {31.06} & {55.80} & 58.80 & 37.10 & 48.71 & 45.69 \\
& TDA \cite{TDA} & 61.35 & 30.29 & 55.54 & {62.58} & {38.12} & 49.58 & 46.63 \\
& \cellcolor{lightgreen} ETTA (Ours) 
& \cellcolor{lightgreen} 61.80 
& \cellcolor{lightgreen} {31.17} 
& \cellcolor{lightgreen} {55.87} 
& \cellcolor{lightgreen} {62.67} 
& \cellcolor{lightgreen} {38.33} 
& \cellcolor{lightgreen} {49.97} 
& \cellcolor{lightgreen} {47.01} \\

\hdashline

\multirow{2}{*}{\footnotesize\rotatebox{90}{\textbf{CSP}}}

& DMN-ZS \cite{DMN} & \textbf{63.87} & {28.57} & {56.12} & {61.44} & \textbf{39.84} & {49.97} & {46.49} \\

& \cellcolor{lightgreen} ETTA\(^+\) (Ours) & \cellcolor{lightgreen} 63.25 & {\cellcolor{lightgreen} {\textbf{31.26}}} & \cellcolor{lightgreen} \textbf{57.05} & \cellcolor{lightgreen} \textbf{62.91} & \cellcolor{lightgreen} {39.23} & \cellcolor{lightgreen} \textbf{50.74} & \cellcolor{lightgreen} \textbf{47.61} \\

\hline
&  &  &  & ViT-B/16 Backbone &  &  & &  \\
\hline
\multirow{2}{*}{\footnotesize\rotatebox{90}{\textbf{SRC}}} 
& \(\text{CLIP}\)-ViT-B/16 \cite{clip} & 66.73 & 47.87 & 60.86 & 73.98 & 46.09 & 59.11 & 57.2 \\
& \(\text{CLIP}^*\)-ViT-B/16 \cite{clip} & 68.34 & 49.89 & 61.88 & 77.65 & 48.24 & 61.20 & 59.42 \\
\hdashline
\multirow{3}{*}{\footnotesize\rotatebox{90}{\textbf{FSA}}} 
& CoOp \cite{coop} & {71.51} & 49.71 & 64.20 & 75.21 & 47.99 & 61.72 & 59.28 \\
& CoCoOp \cite{cocoop} & 71.02 & 50.63 & 64.07 & 76.18 & 48.75 & 62.13 & 59.91 \\
& Tip-Adapter \cite{tip-adapter} & 70.75 & 51.04 & 63.41 & 77.76 & 48.88 & 62.37 & 60.27 \\
\hdashline
\multirow{4}{*}{\footnotesize\rotatebox{90}{\textbf{GP}}} 
& TPT \cite{tpt} & 68.98 & 54.77 & 63.45 & 77.06 & 47.94 & 62.44 & 60.81 \\
& DiffTPT \cite{difftpt} & 70.30 & 55.68 & {65.10} & 75.00 & 46.80 & 62.28 & 60.52 \\
& TDA \cite{TDA} & 69.51 & {60.11} & 64.67 & {80.24} & {50.54} & 65.01 & 63.89 \\
& \cellcolor{lightgreen} ETTA (Ours) 
& \cellcolor{lightgreen} \textcolor{black}{69.58} 
& \cellcolor{lightgreen} {60.87} 
& \cellcolor{lightgreen} {65.26} 
& \cellcolor{lightgreen} {80.42} 
& \cellcolor{lightgreen} {50.98} 
& \cellcolor{lightgreen} {65.42} 
& \cellcolor{lightgreen} {64.38} \\
\hdashline
\multirow{2}{*}{\footnotesize\rotatebox{90}{\textbf{CSP}}} 
& DMN-ZS \cite{DMN} & {72.25} & {58.28} & {65.17} & {78.55} & {53.20} & {65.49} & {63.80} \\
& \cellcolor{lightgreen} ETTA\(^+\) (Ours) 
& \cellcolor{lightgreen} \textbf{72.66} 
& \cellcolor{lightgreen} \textbf{61.18} 
& \cellcolor{lightgreen} \textbf{65.67} 
& \cellcolor{lightgreen} \textbf{80.65} 
& \cellcolor{lightgreen} \textbf{53.63} 
& \cellcolor{lightgreen} \textbf{66.76} 
& \cellcolor{lightgreen} \textbf{65.28} \\

\hline

\end{tabular}
}
\vspace{0.25cm}
\caption{ \small Top-1 Accuracy (\%) Comparison: ETTA vs. Source CLIP (SRC), Few-Shot Adaptation (FSA), TTA with Generic Prompts (GP), and TTA with Class-Specific Prompts (CSP).}
\vspace{-0.4cm}

\label{tab: table1}
\end{table}

\subsection{Implementation Details}

We report top-1 accuracy (\%) using a single NVIDIA A6000 GPU. All models use pre-trained CLIP \cite{clip} with a frozen image encoder (ResNet-50 \cite{resnet} or ViT-B/16 \cite{vit-b16}) and Transformer based text encoder \cite{transformers}. \cite{clip}. Test-time adaptation is performed with a batch size of 1. The filtering strength \(\alpha\) in Adaptive Ensembling is set to 0.6 for OOD and 0.3 for cross-domain datasets, selected via grid search.


\subsection{Results}

\noindent \textbf{Results on OOD Benchmark:} We begin by comparing the performance of the proposed model, ETTA and ETTA\(^+\), with baselines on the OOD benchmark. Tab. \ref{tab: table1} reports results across various OOD datasets, showcasing ETTA’s superior ability to handle distribution shifts. ETTA outperforms all Few-Shot Adaptation (FSA) and Generic Prompt (GP) baselines on both ResNet-50 and ViT-B/16 backbones, achieving an average accuracy improvement of \textbf{3.12\%} and \textbf{3.57\%} over TPT \cite{tpt}, respectively. Additionally, ETTA surpasses DiffTPT \cite{difftpt} and TDA \cite{TDA}, with gains of \textbf{1.32\%} and \textbf{0.38\%} on ResNet-50 and \textbf{3.86\%} and \textbf{0.49\%} on ViT-B/16. Further improvements are observed with Class-Specific Prompts (CSP), where ETTA\(^+\) achieves the highest accuracy, outperforming its competing approach, DMN-ZS, by \textbf{1.12\%} on ResNet-50 and \textbf{1.48\%} on ViT-B/16.

\vspace{.2cm}

\begin{table}[!t]
\centering
\renewcommand{\arraystretch}{1}  
\setlength{\tabcolsep}{4pt}
\resizebox{0.84\linewidth}{!}{
\begin{tabular}{clcccccccccccc}
\hline
\rowcolor{verylightgray}
& \textbf{Method} & \textbf{Aircraft} & \textbf{Caltech101} & \textbf{Cars} & \textbf{DTD} & \textbf{EuroSAT} & \textbf{Flower102} & \textbf{Food101} & \textbf{Pets} & \textbf{SUN397} & \textbf{UCF101} & \textbf{Average} \\
\hline
& & & & & & ResNet50 Backbone  & & & & & &\\
\hline
\multirow{2}{*}{\footnotesize\rotatebox{90}{\textbf{SRC}}}  
& CLIP-ResNet50 \cite{clip}   & 15.66 & 85.88 & 55.70 & 40.37 & 23.69 & 61.75 & 73.97 & 83.57 & 58.80 & 58.84 & 55.82 \\
& \(\text{CLIP}^*\text{-ResNet50}\) \cite{clip} & 16.11 & 87.26 & 55.89 & 40.37 & 25.79 & 62.77 & 74.82 & 82.97 & 60.85 & 59.48 & 56.63 \\
\hdashline

\multirow{2}{*}{\footnotesize\rotatebox{90}{\textbf{FSA}}}  
& CoOp \cite{coop} & 15.12 & 86.53 & 55.32 & 37.29 & 26.20 & 61.55 & 75.59 & 87.00 & 58.15 & 59.05 & 56.18 \\
& CoCoOp \cite{cocoop} & 14.61 & 87.38 & 56.22 & 38.53 & 28.73 & 65.57 & 76.20 & {88.39} & 59.61 & 57.10 & 57.23 \\
\hdashline

\multirow{4}{*}{\footnotesize\rotatebox{90}{\textbf{GP}}}  
& TPT \cite{tpt} & 17.58 & 87.02 & 58.46 & 40.84 & 28.33 & 62.69 & 74.88 & 84.49 & 61.46 & 60.82 & 57.66 \\
& DiffTPT \cite{difftpt} & 17.60 & 86.89 & \textbf{60.71} & 40.72 & {41.04} & 63.53 & \textbf{79.21} & 83.40 & {62.72} & 62.67 & 59.85 \\
& TDA \cite{TDA} & \textcolor{black}{17.61} & \textcolor{black}{89.70} & \textcolor{black}{57.78} & \textcolor{black}{43.74} & \textcolor{black}{42.11} & \textcolor{black}{68.74} & \textcolor{black}{77.75} & \textcolor{black}{86.18} & \textcolor{black}{62.53} & \textcolor{black}{64.18} & \textcolor{black}{61.03} \\

& \cellcolor{lightgreen} ETTA (Ours)  
& \cellcolor{lightgreen} \textcolor{black}{18.84}  
& \cellcolor{lightgreen} \textcolor{black}{90.18}  
& \cellcolor{lightgreen} \textcolor{black}{58.60}  
& \cellcolor{lightgreen} \textcolor{black}{42.73}  
& \cellcolor{lightgreen} \textcolor{black}{46.01}  
& \cellcolor{lightgreen} \textcolor{black}{69.75}  
& \cellcolor{lightgreen} \textcolor{black}{78.05}  
& \cellcolor{lightgreen} \textcolor{black}{86.42}  
& \cellcolor{lightgreen} \textcolor{black}{{62.79}}  
& \cellcolor{lightgreen} \textcolor{black}{{65.48}}  
& \cellcolor{lightgreen} \textcolor{black}{{61.88}} \\

\hdashline

\multirow{4}{*}{\footnotesize\rotatebox{90}{\textbf{CSP}}}
& CuPL \cite{CuPL} & \textcolor{black}{19.59} & \textcolor{black}{89.29} & \textcolor{black}{57.28} & {48.64} & \textcolor{black}{38.38} & \textcolor{black}{65.44} & \textcolor{black}{ 76.94} & \textcolor{black}{84.84} & \textcolor{black}{62.55} & \textcolor{black}{58.97} & \textcolor{black}{ 60.31} \\

& SuS-X \cite{Sus-x} & \textcolor{black}{19.47} & \textcolor{black}{89.53} & \textcolor{black}{57.27} & {50.59} & \textcolor{black}{45.57} & \textcolor{black}{67.72} & \textcolor{black}{77.58} & \textcolor{black}{85.34} & \textcolor{black}{62.95} & \textcolor{black}{61.54} & \textcolor{black}{61.76} \\

& DMN-ZS \cite{DMN} & \textcolor{black}{22.77} & \textcolor{black}{90.14} & \textcolor{black}{60.02} & {50.41} & \textcolor{black}{48.72} & \textcolor{black}{67.93} & \textcolor{black}{76.70} & \textcolor{black}{86.78} & \textcolor{black}{\textbf{64.39}} & \textcolor{black}{65.34} & \textcolor{black}{63.71} \\

& \cellcolor{lightgreen} ETTA\(^+\) (Ours)  
& \cellcolor{lightgreen} \textcolor{black}{\textbf{23.34}}  
& \cellcolor{lightgreen} \textcolor{black}{\textbf{90.67}}  
& \cellcolor{lightgreen} \textcolor{black}{{60.24}}  
& \cellcolor{lightgreen} \textcolor{black}{\textbf{51.83}}  
& \cellcolor{lightgreen} \textcolor{black}{\textbf{52.30}}  
& \cellcolor{lightgreen} \textcolor{black}{\textbf{69.92}}  
& \cellcolor{lightgreen} \textcolor{black}{78.17}  
& \cellcolor{lightgreen} \textcolor{black}{\textbf{88.48}}  
& \cellcolor{lightgreen} \textcolor{black}{{63.93}}  
& \cellcolor{lightgreen} \textcolor{black}{\textbf{65.55}}  
& \cellcolor{lightgreen} \textcolor{black}{\textbf{64.44}} \\

\hline
& & & & & & ViT-B/16 Backbone  & & & & & &\\
\hline

\multirow{2}{*}{\footnotesize\rotatebox{90}{\textbf{SRC}}}  
& CLIP-ViT-B/16 \cite{clip} & 23.67 & 93.35 & 65.48 & 44.27 & 42.01 & 67.44 & 83.65 & 88.25 & 62.59 & 65.13 & 63.58 \\
& \(\text{CLIP}^*\text{-ViT-B/16}\) \cite{clip} & 23.22 & 93.55 & 66.11 & 45.04 & 50.42 & 66.99 & 82.86 & 86.92 & 65.63 & 65.16 & 64.59 \\
\hdashline

\multirow{2}{*}{\footnotesize\rotatebox{90}{\textbf{FSA}}}  
& CoOp \cite{coop} & 18.47 & 93.70 & 64.51 & 41.92 & 46.39 & 68.71 & 85.30 & 89.14 & 64.15 & 66.55 & 63.88 \\
& CoCoOp \cite{cocoop} & 22.29 & 93.79 & 64.90 & 45.45 & 39.23 & 70.85 & 83.97 & {90.46} & 66.89 & 68.44 & 64.63 \\
\hdashline

\multirow{4}{*}{\footnotesize\rotatebox{90}{\textbf{GP}}}  
& TPT \cite{tpt}& 24.78 & {94.16} & 66.87 & {47.75} & 42.44 & 68.98 & 84.67 & 87.79 & 65.50 & 68.04 & 65.10 \\
& DiffTPT \cite{difftpt} & {25.60} & 92.49 & 67.01 & 47.00 & 43.13 & 70.10 & {87.23} & 88.22 & 65.74 & 62.67 & 65.47 \\

& TDA \cite{TDA} & \textcolor{black}{{23.91}} & \textcolor{black}{94.24} & \textcolor{black}{ 67.28} & \textcolor{black}{ 47.40} & \textcolor{black}{{ 58.00}} & \textcolor{black}{ 71.42} & \textcolor{black}{{86.14}} & \textcolor{black}{88.63} & \textcolor{black}{{67.62}} & \textcolor{black}{{70.66}} & \textcolor{black}{67.53} \\
&\cellcolor{lightgreen} ETTA (Ours) & \cellcolor{lightgreen} \textcolor{black}{{25.65}} & \cellcolor{lightgreen} \textcolor{black}{{94.74}} & \cellcolor{lightgreen} \textcolor{black}{{68.40}} & \cellcolor{lightgreen} {{49.64}} & \cellcolor{lightgreen} \textcolor{black}{\textbf{59.86}} & \cellcolor{lightgreen} \textcolor{black}{{73.86}} & \cellcolor{lightgreen} \textcolor{black}{86.61} & \cellcolor{lightgreen} \textcolor{black}{{90.08}} & \cellcolor{lightgreen} \textcolor{black}{{68.47}}&  \cellcolor{lightgreen} \textcolor{black}{{71.95}} & \cellcolor{lightgreen} \textcolor{black}{{68.93}} \\

\hdashline

\multirow{2}{*}{\footnotesize\rotatebox{90}{\textbf{CSP}}}  
& DMN-ZS \cite{DMN}  & \textcolor{black}{{30.03}} & \textbf{95.38} & \textcolor{black}{{67.96}} & {{55.85}} & \textcolor{black}{{59.43}} & \textcolor{black}{{74.49}} & \textcolor{black}{85.08} & \textcolor{black}{{92.04}} & \textcolor{black}{{70.18}}& \textcolor{black}{{72.51}} & \textcolor{black}{{70.72}}\\

& \cellcolor{lightgreen}ETTA\(^+\) (Ours) & \cellcolor{lightgreen} \textcolor{black}{\textbf{30.18}} & \cellcolor{lightgreen} \textcolor{black}{{95.25}} & \cellcolor{lightgreen} \textcolor{black}{\textbf{68.51}} & \cellcolor{lightgreen} \textbf{{56.20}} & \cellcolor{lightgreen} \textcolor{black}{{58.86}} & \cellcolor{lightgreen} \textcolor{black}{\textbf{77.43}} & \cellcolor{lightgreen} \textcolor{black}{\textbf{87.38}} & \cellcolor{lightgreen} \textcolor{black}{\textbf{92.15}} & \cellcolor{lightgreen} \textcolor{black}{\textbf{71.11}}& \cellcolor{lightgreen} \textcolor{black}{\textbf{73.78}} & \cellcolor{lightgreen} \textcolor{black}{\textbf{71.09}} \\
\hline
\end{tabular}
}
\vspace{0.25cm}
\caption{\small Top-1 Accuracy (\%) Comparison: ETTA vs. Source CLIP (SRC), Few-Shot Adaptation (FSA), TTA with Generic Prompts (GP), and TTA with Class-Specific Prompts (CSP).}
 \vspace{-.5cm}
\label{tab: table2}
\end{table}

\noindent \textbf{Results on Cross-Domain Benchmark:} 
We further evaluate ETTA’s cross-domain adaptability by comparing it with baselines on the cross-domain benchmark. As shown in Tab. \ref{tab: table2}, ETTA consistently outperforms prompt-tuning-based TTA models, including TPT \cite{tpt} and DiffTPT \cite{difftpt}, as well as cache-based TDA \cite{TDA}, achieving the highest average accuracy across the FSA and GP categories. On ResNet-50, ETTA surpasses TPT by \textbf{4.22\%}, DiffTPT by \textbf{2.03\%}, and TDA by \textbf{0.85\%}. On ViT-B/16, the gains are even more pronounced: \textbf{3.83\%} over TPT, \textbf{3.46\%} over DiffTPT, and \textbf{1.4\%} over TDA. These improvements are especially notable on datasets like EuroSAT \cite{eurosat}, Flower102 \cite{flower}, and SUN397 \cite{sun397}, which span diverse visual domains.  
For Class-Specific Prompts (CSP), we evaluate ETTA\(^+\) against CuPL \cite{CuPL}, SuS-X \cite{Sus-x}, and DMN \cite{DMN} on ResNet-50, and against DMN on ViT-B/16. ETTA\(^+\) achieves the highest average accuracy, outperforming all state-of-the-art methods.

\begin{wraptable}{r}{0.5\textwidth}
\centering
\scriptsize
\setlength{\tabcolsep}{4pt}
\renewcommand{\arraystretch}{1.1}
\begin{tabular}{lccccc}
\hline
\rowcolor{verylightgray}
\textbf{Method} & \(\textbf{CLIP}^*\) & \textbf{TPT} & \textbf{DiffTPT} & \textbf{TDA} & \textbf{ETTA} \\
\hline
\textbf{Time} & \textbf{18m} & 19h 30m & 51h 15m & 25m & \underline{\textbf{19m}} \\
\textbf{Acc.} & 59.81 & 60.74 & 60.80 & 61.35 & \underline{\textbf{61.80}} \\
\hline
\end{tabular}
\vspace{0.3cm}
\caption{\small Test-time comparison with on ImageNet.}
\label{tab:3}
\end{wraptable}

\vspace{.2cm}
\noindent \textbf{Time Efficiency Comparison: } We evaluated ETTA’s efficiency on ImageNet \cite{imagenet} using ResNet-50 \cite{resnet} on an NVIDIA A6000 GPU. Tab. \ref{tab:3} compares its accuracy and testing time with TPT \cite{tpt}, DiffTPT \cite{difftpt}, and TDA \cite{TDA}. ETTA processes 50,000 images in 19 minutes (18 min for zero-shot CLIP + 1 min overhead), versus 25 min for TDA and 19 and 51 hours for tuning-based TPT and DiffTPT, respectively. These results show ETTA’s high accuracy with minimal overhead, making it ideal for real-time use.

\subsection{Ablation Studies}
\label{sec:ablation}

\noindent{\textbf{Why Mimicking an Unbounded Cache? }}

\begin{wrapfigure}{r}{0.5\textwidth}
    \vspace{-1.2cm}
    \centering
    \includegraphics[width=0.99\linewidth]{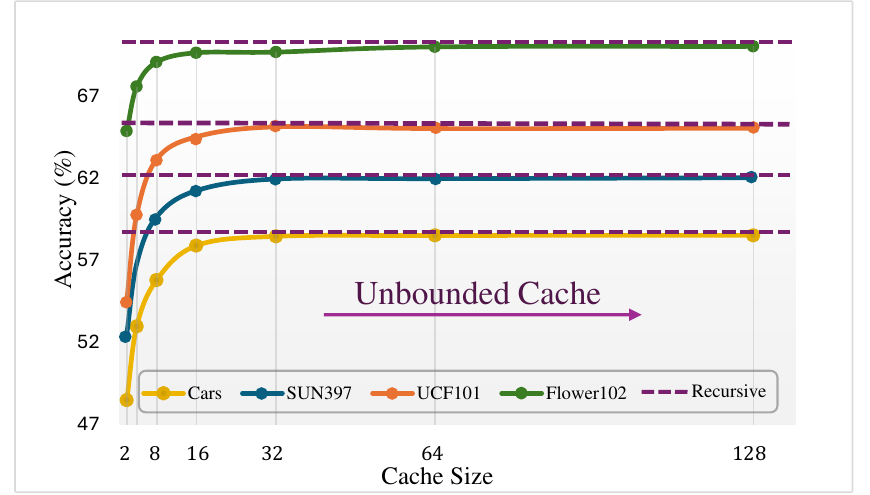}
    \vspace{-5pt}
    \caption{\small ETTA accuracy in the bounded cache setup with varying cache sizes on cross-domain datasets (ResNet-50 backbone). Performance improves with larger cache sizes, approaching the accuracy of ETTA's Recursive Update cache.}
    \label{fig:clip-ablation-unbounded}
    \vspace{-10pt}
\end{wrapfigure}

\noindent Our proposed ETTA framework is equivalent to an unbounded cache, allowing all test images to contribute to the decision boundary. A key question is whether more cache prototypes always improve performance in the cross-attention setting. To explore this, we remove Recursive Update (Eqs. \ref{eq:recursive1}–\ref{eq:recursive3}) and instead use finite caches of varying sizes, computing contextual embeddings via Eq. \ref{eq:(4)}. Fig. \ref{fig:clip-ablation-unbounded} shows accuracy in the bounded setting across cross-domain datasets. Performance improves with cache size, converging to ETTA's Recursive Update cache, showing our method reaches the upper bound of bounded cache performance—without storing test samples. More results are available in appendix \ref{sec:robustness}.

\begin{figure}[tbp]
  \centering

  \begin{minipage}{0.48\textwidth}
    \centering
    \includegraphics[width=\linewidth]{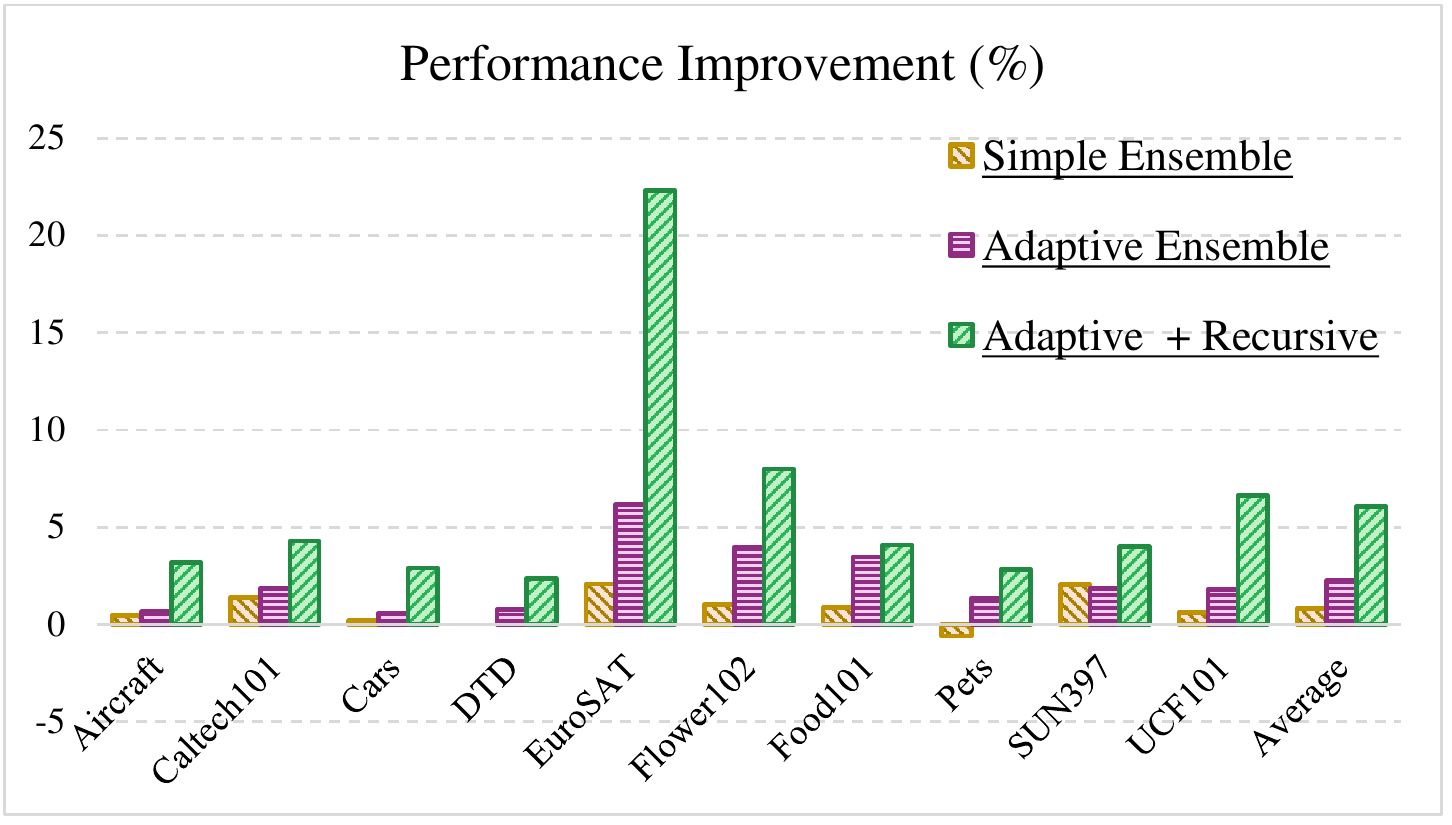}
  \end{minipage}
  \hfill
  \begin{minipage}{0.48\textwidth}
    \centering
    \includegraphics[width=\linewidth]{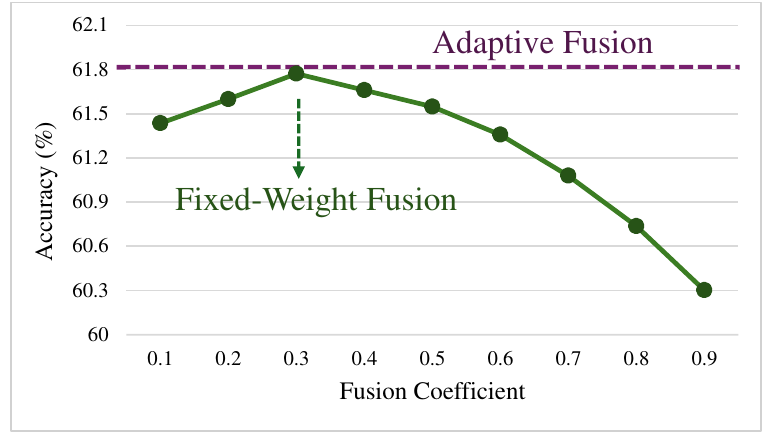}
  \end{minipage}
\vspace{0.2cm}
  \caption{\small \textbf{Left:} Accuracy gains across cross-domain datasets using our Adaptive Ensembling, compared to Simple Ensembling \cite{clip} and its combination with Recursive Update. \textbf{Right:} Accuracy of the ETTA method on the ResNet-50 backbone, comparing fixed-weight fusion to adaptive fusion.}
  \label{fig:4}
\end{figure}

\vspace{.2cm}

\noindent{\textbf{Effectiveness of Adaptive Ensemble Module:}} The Adaptive Ensemble module in ETTA filters out prompts with limited class-relevant information based on the input image embedding. Accuracy gain (defined as the difference between the accuracy of each model and the standard zero-shot \(\text{CLIP}^*\)) is used to evaluate its effectiveness on cross-domain datasets. As shown in Fig. \ref{fig:4} (Left), Adaptive Ensembling consistently outperforms Simple Ensembling, proposed by Radford et al. \cite{clip}. The figure also illustrates the combined contributions of Adaptive Ensembling and Recursive Updating, highlighting their complementary roles in ETTA. Additional experiments are provided in the supplementary section \ref{sec:Adaptive-Ensemble}.

\vspace{0.2cm}
\noindent{\textbf{Impact of Prompt Filtering Coefficient:}} We conduct parameter studies on the prompt filtering coefficient (\(\alpha\)), which controls the proportion of prompts filtered before average in the Adaptive Ensembling module.

\begin{wraptable}{r}{0.6\textwidth}
\vspace{-0.6cm}
\centering
\tiny  
\setlength{\tabcolsep}{4pt}
\renewcommand{\arraystretch}{2}
\begin{tabular}{l|cccccccccc}
\hline
\rowcolor{verylightgray}
\textbf{\(\alpha\)}  & 0.1 & 0.2 & 0.3 & 0.4 & 0.5 & 0.6 & 0.7 & 0.8 & 0.9 & 1.0 \\
\hline
\textbf{Acc (\%)} & 59.25 & 59.65 & 59.84 & 59.85 & 59.90 & \textbf{60.02} & 59.93 & 59.93 & 59.90 & 59.81 \\
\textbf{Acc (\%)} & 59.12 & 59.22 & \textbf{59.26} & 59.15 & 59.02 & 58.82 & 58.62 & 58.61 & 58.45 & 58.20 \\
\end{tabular}
\vspace{0.3cm}
\caption{\small Accuracy of Adaptive Ensembling across \(\alpha\) values on ImageNet (top) and cross-domain datasets (bottom).}
\label{tab:alpha-ablation}
\vspace{-10pt}
\end{wraptable}

As shown in Tab. \ref{tab:alpha-ablation}, ETTA's performance varies across filtering strengths, with results on ImageNet \cite{imagenet} validation set in the first row and cross-domain averages in the second row. Optimal performance is achieved with \(\alpha = 0.6\) for ImageNet, and \(\alpha = 0.3\) for cross-domain datasets. A lower \(\alpha\) filters out more irrelevant prompts, beneficial in distinct domains with few highly relevant prompts.

\vspace{0.2cm}
\noindent{\textbf{Effectiveness of the Adaptive Logit Fusion:}} Adaptive Fusion, the final component of ETTA, combines logits from the Adaptive Ensemble and Recursive Update modules, weighting them by confidence to prioritize reliable predictions. We compare it with fixed-weight fusion, which combines logits using constant weights \(1 - \beta\) and \(\beta\). As shown in Fig.~\ref{fig:4} (Right), fixed-weight fusion peaks at \( \beta = 0.3 \) on ImageNet \cite{imagenet}, closely matching Adaptive Fusion. However, Adaptive Fusion requires no dataset-specific tuning, making it more robust and generalizable. Additional cross-domain results in the supplementary (Section~\ref{sec:Adaptive-logits}).

\section{Conclusion}

We propose ETTA, a training-free, efficient test-time adaptation method for vision-language models like CLIP \cite{clip}. ETTA introduces two key components: an Adaptive Ensemble module that selects relevant prompts per test sample, and a Recursive Update module that incrementally refines the decision boundary using test images. This strategy matches the performance of an unbounded cache while keeping memory and computation low. An Adaptive Fusion block combines logits from both modules using confidence scores to exploit their complementary strengths. ETTA achieves state-of-the-art results across benchmarks, making it practical for real-world use. Future work includes adding augmentations to Recursive Updating to enhance decision boundaries and applying ETTA to few-shot settings with limited labeled data.

\section{Acknowledgment}
This work was supported by the Natural Sciences and Engineering Research Council of Canada (NSERC) through the Smart Autonomous Robotic Technology for Space Exploration (SMART-ART) CREATE program.
\newpage
{
    \small
    \bibliography{main}
}

\newpage
\appendix

\section{Dataset Details}
\label{sec:datasets}
This section provides further details about the datasets used to evaluate the proposed TTA method. Out-of-distribution (OOD) datasets examine the model's robustness against distribution shifts, while cross-domain datasets assess its effectiveness across varying domains.

\vspace{-0.2cm}
\subsection{OOD Datasets}
\vspace{0.1cm}
\begin{itemize}
    \item ImageNet-A \cite{imagenet-a}: This dataset comprises 7,500 images spanning 200 challenging classes, selected as a subset of the original ImageNet \cite{imagenet}. These naturally adversarial samples feature real-world complexities like unusual lighting, textures, poses, and occlusions. Testing on ImageNet-A is essential for identifying model weaknesses in handling complex, unpredictable data that differs from typical training conditions.
    \item ImageNet-V2 \cite{imagenet-v2}: This dataset includes 10,000 images across 1,000 classes, aligned with the original ImageNet \cite{imagenet} categories but with new, distinct images. By closely resembling the original set without exact duplication, ImageNet-V2 helps reveal whether a model is overfitted to the specific instances in the original ImageNet, providing a clearer view of its true generalization ability.
    \item ImageNet-R \cite{imagenet-r}: This dataset contains 30,000 images across 200 classes, featuring objects in non-traditional, artistic styles like cartoons, paintings, sculptures, and embroidery. It serves as a test set for evaluating model robustness to extreme stylistic shifts that significantly transform an object’s appearance.
    \item ImageNet-S \cite{imagenet-s}: This dataset includes 50,000 images spanning 1,000 classes, consistent with the original ImageNet \cite{imagenet} categories. It serves to test model robustness against various artistic styles, such as sketches and abstract depictions. This evaluation examines whether models can generalize object recognition without relying on realistic textures and adapt to representations with modified or minimal texture details.    
\end{itemize}
\vspace{-0.5cm}
\subsection{Cross-Domain Datasets}
Cross-domain datasets assess model robustness across tasks like fine-grained classification, scene, and texture recognition. Each provides a test set for TTA evaluation and a validation set for hyperparameter tuning.

\begin{itemize}
    \item Aircraft \cite{aircraft}: Contains 3,333 test images across 100 classes, challenging models to distinguish fine-grained variations between aircraft types. We use the remaining 6,667 images as the validation set.
    \item Caltech101 \cite{caltech}: Comprises 2,465 test images across 101 object classes, evaluating models on their ability to recognize diverse objects. The remaining 6,681 images are used as the validation set.
    \item Cars \cite{cars}: Includes 8,041 test images across 196 car models, focusing on fine-grained distinctions between different types. A remaining set of 8,144 images is also provided but excluded from TTA evaluations.
    \item DTD \cite{dtd}: Contains 1,692 test images across 47 texture classes, designed to assess texture recognition capabilities. The remaining 3,948 images are used for validation in this study.
    \item EuroSAT \cite{eurosat}: Comprises 8,100 test images across 10 satellite image classes, evaluating performance in remote sensing and geographic classification. The dataset includes 18,900 more images which are excluded from the TTA evaluation.
    \item Flower102 \cite{flower}: Contains 2,463 test images across 102 flower classes, offering a challenging set for fine-grained floral classification. The remaining images are used as the validation set for hyperparameter tuning.
    \item Food101 \cite{food}: Features 30,300 test images across 101 food classes, assessing a model’s ability to recognize diverse dishes. Additionally, the dataset includes a training set of 70,700 images.
    \item Pets \cite{pets}: Includes 3,669 test images across 37 pet breed classes, challenging models with fine-grained classification of cats and dogs. The dataset includes an additional set of 3,680 images, which is not utilized in TTA evaluations.
    \item SUN397 \cite{sun397}: This dataset comprises 108,754 images spanning 397 scene categories, designed to evaluate a model's capacity to classify a diverse array of indoor and outdoor environments. Of these, 19,850 images are allocated to the test set, leaving the remainder for training or validation.
    \item UCF101 \cite{ucf101}: Contains 3,783 test images across 101 action classes, testing model performance on human action classification. The dataset includes a training set of 9,537 images, offering additional data for other experimental uses.
\end{itemize}

\label{sec:rationale}
\vspace{-0.5cm}
\section{ETTA Algorithm}
\label{sec:algorithm}
\vspace{-0.2cm}
\begin{algorithm}[H]
\scriptsize
\caption{The proposed ETTA algorithm.}
\begin{algorithmic}[1]
    \State \textbf{Input:} Input image $\mathbf{I}$, encoder $\operatorname{E_v}(.)$, normalized text embeddings $\{\mathbf{w}_i^{(k)}\}$ for $i = 1, \dots, C$, $k = 1, \dots, T$.
    \State \textbf{Initialize:} $s_i^{\text{old}} = 0$ and $\widehat{\mathbf{w}}_i^{\text{old}} = 0$ for each $i$.
    \For{all arriving images}
        \State $\mathbf{v}_n \gets \frac{\operatorname{E_v}(\mathbf{I})}{\| \operatorname{E_v}(\mathbf{I})\|}$ \hfill (normalized image embedding)
        \State $\mathcal{W}_i^{\alpha} = \left\{ \mathbf{w}_i^{(k)} : \langle \mathbf{w}_i^{(k)}, \mathbf{v} \rangle \geq \gamma_{\alpha} \right\}$ \hfill (filter top-$\alpha$)
        \State $\overline{\mathbf{w}}_i = \frac{\sum_{\mathbf{w}_i^{(k)} \in \mathcal{W}_i^{\alpha}} \mathbf{w}_i^{(k)}}{\left\|\sum_{\mathbf{w}_i^{(k)} \in \mathcal{W}_i^{\alpha}} \mathbf{w}_i^{(k)}\right\|}$ \hfill (average \& normalize)
        \State $l_i^{(a)} \gets \langle \overline{\mathbf{w}}_i, \mathbf{v} \rangle, \quad \forall i$ \hfill \textcolor{purple}{(adaptive ensemble logits)}
        \State $\hat{i} \gets \arg\max_i \{l_i^{(a)}\}$ \hfill (pseudo-label)
        \State $\widehat{\mathbf{w}}_{\hat{i}}^{\text{new}} \gets \frac{s_{\hat{i}}^{\text{old}} \, \widehat{\mathbf{w}}_{\hat{i}}^{\text{old}} + \exp(\langle \overline{\mathbf{w}}_{\hat{i}}, \mathbf{v} \rangle) \mathbf{v}}{s_{\hat{i}}^{\text{old}} + \exp(\langle \overline{\mathbf{w}}_{\hat{i}}, \mathbf{v} \rangle)}$
        \State $s_{\hat{i}}^{\text{new}} \gets s_{\hat{i}}^{\text{old}} + \exp(\langle \overline{\mathbf{w}}_{\hat{i}}, \mathbf{v} \rangle)$
        \State $\widehat{\mathbf{w}}_i^{\text{new}} \gets \widehat{\mathbf{w}}_i^{\text{old}}, \quad s_i^{\text{new}} \gets s_i^{\text{old}}, \quad \forall i \neq \hat{i}$
        \State $l_i^{(r)} \gets \frac{\langle \widehat{\mathbf{w}}_i^{\text{new}}, \mathbf{v} \rangle}{\|\widehat{\mathbf{w}}_i^{\text{new}}\|}, \quad \forall i$ \hfill \textcolor{green!50!black}{(recursive logits)}
        \State $l_i^{(c)} \gets \frac{H^{(r)}}{H^{(a)} + H^{(r)}} l_i^{(a)} + \frac{H^{(a)}}{H^{(a)} + H^{(r)}} l_i^{(r)}$ \hfill \textcolor{orange!80!black}{(fusion)}
        \State $\hat{y} \gets \arg\max_i \{l_i^{(c)}\}$ \hfill (final prediction)
    \EndFor
\end{algorithmic}
\end{algorithm}

\section{More Experimental Results}
\subsection{Effectiveness of Adaptive Ensemble Module}
\label{sec:Adaptive-Ensemble}

\begin{figure}[!t]
    \centering
    \includegraphics[width=0.84\linewidth]{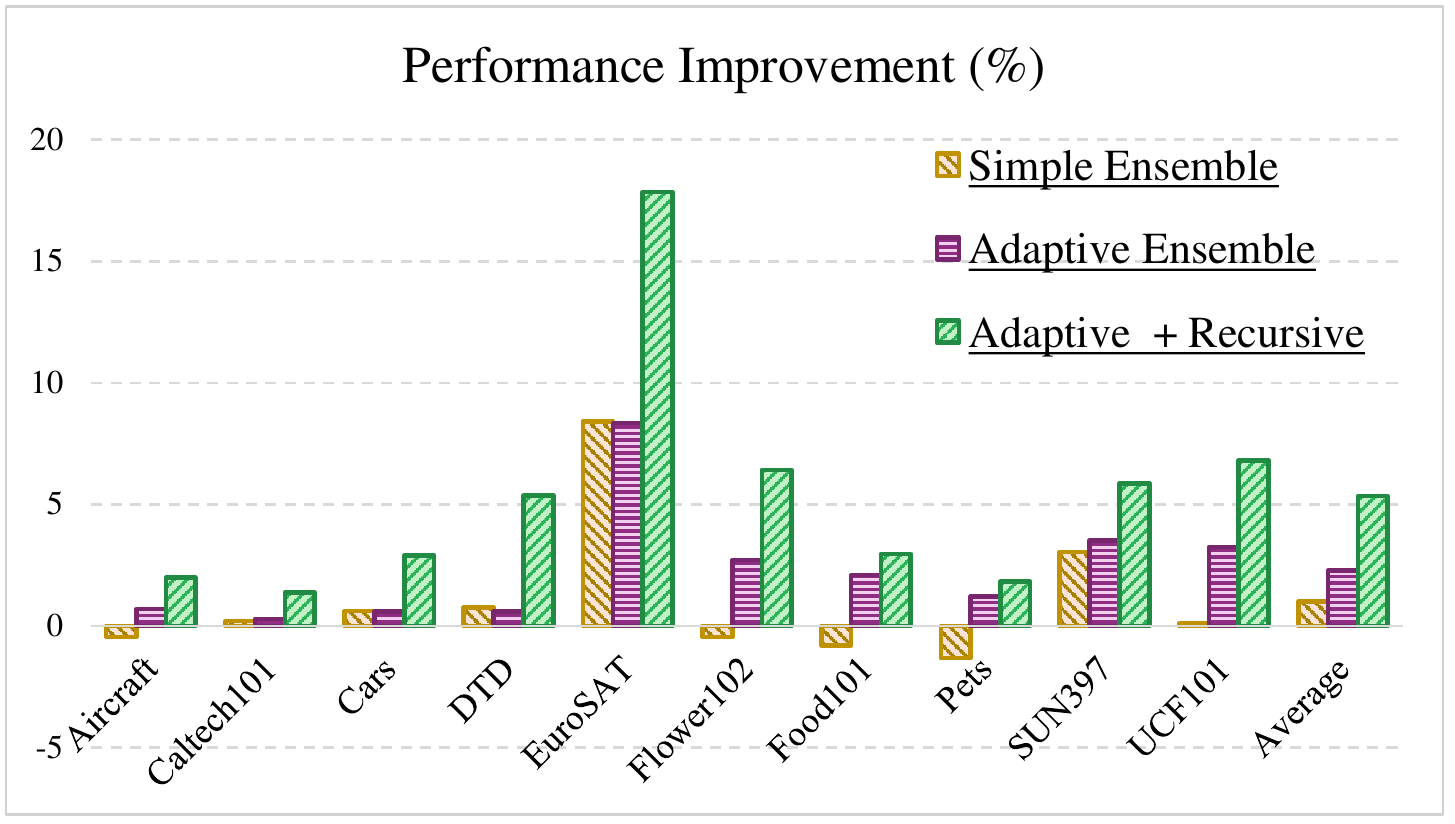}
    \vspace{0.2cm}
    \caption{\small Accuracy improvement across cross-domain datasets for the Adaptive Ensembling method versus Simple Ensemble approach. Network architecture: ViT-B/16.}
    \label{fig:adaptive-vs-simple-ViT-B}
\end{figure}

\begin{figure}[!t]
    \centering
    \includegraphics[width=0.84\linewidth]{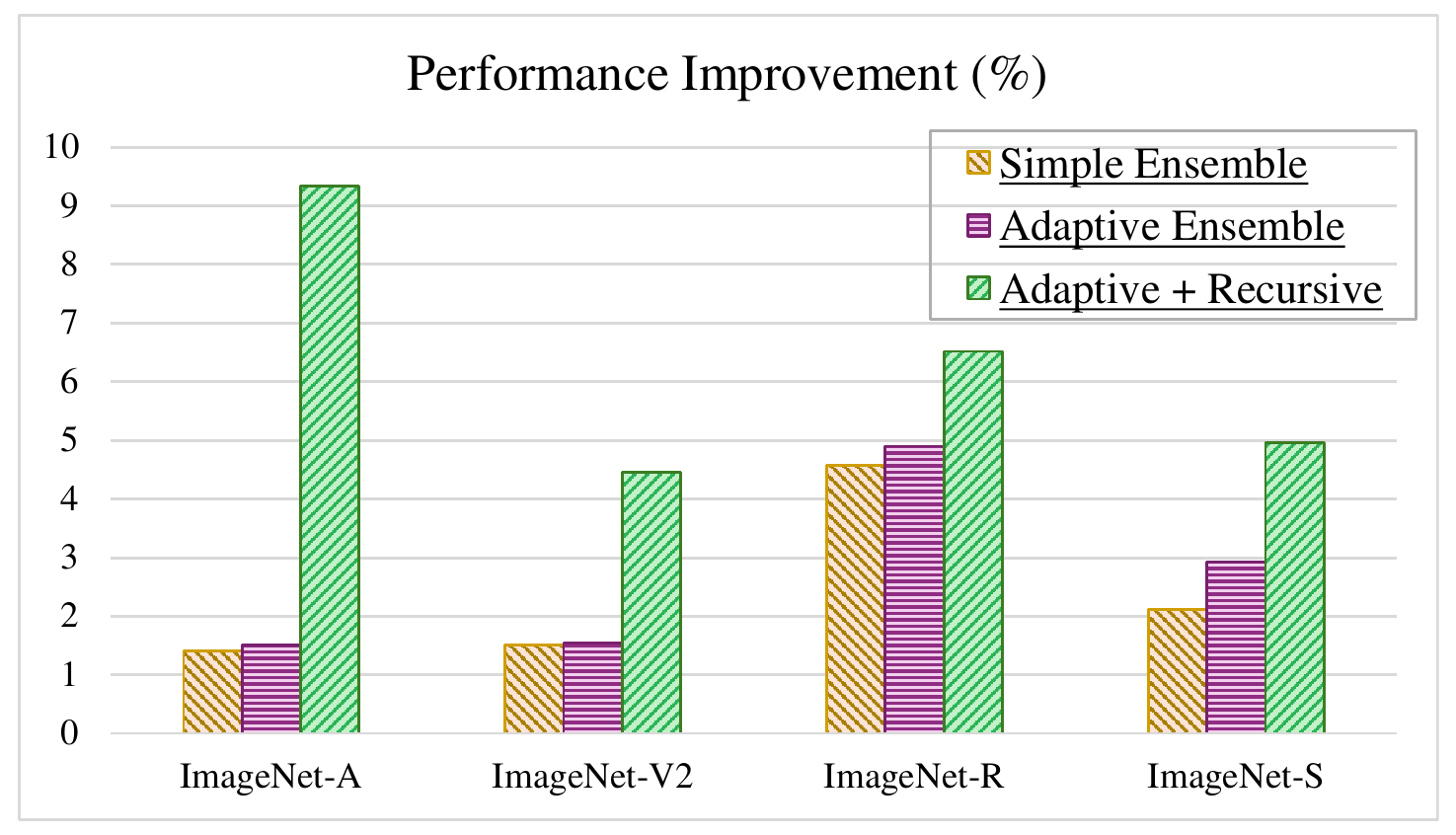}
    \vspace{0.2cm}
    \caption{\small Accuracy improvement across OOD datasets for the Adaptive Ensembling method versus Simple Ensemble approach. Network architecture: ResNet-50.}
    \label{fig:adaptive-vs-simple-ViT-ood}
\end{figure}

In Section \ref{sec:ablation}, we demonstrated the effectiveness of the Adaptive Ensemble module in improving accuracy on cross-domain datasets with the ResNet-50 backbone \cite{resnet}. To further validate its robustness, we present additional results using the ViT-B/16 backbone \cite{vit-b16} in Fig. \ref{fig:adaptive-vs-simple-ViT-B}. The figure highlights that Adaptive Ensembling significantly outperforms Simple Ensembling, which was proposed by Radford et al. \cite{clip}. Notably, Simple Ensembling fails to improve performance on some datasets, such as Flower102 \cite{flower}, Food101 \cite{food}, and Pets \cite{pets}, where it even leads to a decline in accuracy. In contrast, our Adaptive Ensemble strategy consistently enhances performance across all datasets. The figure also demonstrates the impact of combining the Adaptive Ensemble module with the Recursive Update module, as shown by the green bars. Together, these modules deliver complementary benefits, resulting in significant and consistent accuracy gains. Furthermore, we present a comparison of performance gains between the Simple Ensemble and Adaptive Ensemble methods across OOD datasets using the ResNet-50 backbone in Fig. \ref{fig:adaptive-vs-simple-ViT-ood}. Similar to previous observations, the Adaptive Ensembling strategy achieves greater accuracy improvements than Simple Ensembling on OOD datasets. Additionally, the combination of Adaptive Ensembling and Recursive Update modules delivers the best performance, highlighting their complementary strengths.

\subsection{Prompt Filtering Coefficient}
\label{sec:alpha}
In Eq.~\ref{eq:filter}, the hyperparameter \(\alpha\) controls the percentage of prompts retained for each class before ensembling. In Section~\ref{sec:ablation}, we conducted a parameter study for \(\alpha\) on OOD datasets using the ImageNet validation set \cite{imagenet} and on cross-domain datasets using their non-test sets as validation. The reported results, averaged over all cross-domain datasets, show that the optimal value of \(\alpha\) is 0.6 for OOD datasets and 0.3 for cross-domain datasets.
\begin{wraptable}{r}{0.6\linewidth}
\centering
\tiny
\setlength{\tabcolsep}{4pt}
\renewcommand{\arraystretch}{1.2}
\begin{tabular}{l|cccccccccc}
\rowcolor{verylightgray}
\hline
\textbf{\(\alpha\)}  & 0.1 & 0.2 & 0.3 & 0.4 & 0.5 & 0.6 & 0.7 & 0.8 & 0.9 & 1.0 \\
\hline
\textbf{Acc (\%)} & 67.76 & 68.04 & 68.08 & 68.22 & 68.31 & \textbf{68.42} & 68.38 & 68.40 & 68.36 & 68.34 \\
\textbf{Acc (\%)} & 65.86 & 66.23 & \textbf{66.24} & 66.15 & 66.12 & 65.85 & 65.92 & 65.72 & 65.64 & 65.50 \\
\end{tabular}
\vspace{0.2cm}
\caption{\small Accuracy of Adaptive Ensembling across \(\alpha\) values on ImageNet (top) and cross-domain datasets (bottom), using ViT-B/16.}
\label{tab:table5}
\vspace{-10pt}
\end{wraptable}
To ensure the consistency of these findings across backbones, we repeated the experiments using the ViT-B/16 backbone. As shown in Tab. \ref{tab:table5}, the results confirm that 0.6 remains optimal for OOD datasets, while 0.3 works best for cross-domain datasets, indicating that cross-domain datasets require filtering a larger proportion of prompts to achieve optimal performance.

\subsection{Effectiveness of Adaptive Logit Fusion}
\label{sec:Adaptive-logits}

\begin{figure}[!h]
    \centering
    \setlength{\tabcolsep}{2pt} 
    \renewcommand{\arraystretch}{1} 

    \begin{tabular}{cc}
        \begin{minipage}{0.4\textwidth}
            \centering
            \includegraphics[width=\linewidth]{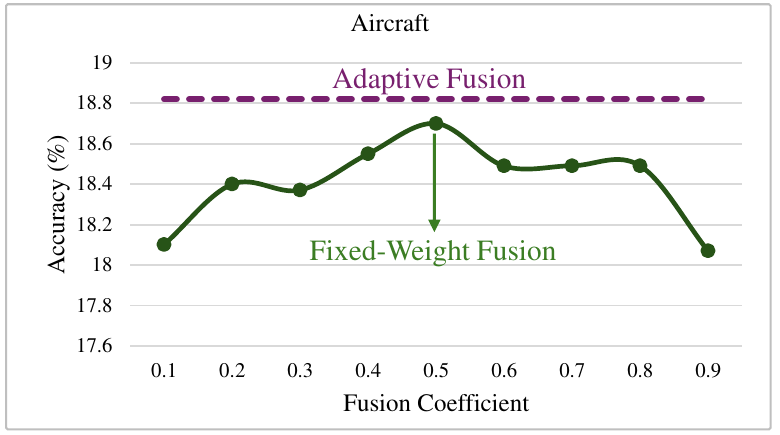}
            \vspace{0.1cm}
            \caption*{(a) Aircraft}
        \end{minipage} &
        \begin{minipage}{0.4\textwidth}
            \centering
            \includegraphics[width=\linewidth]{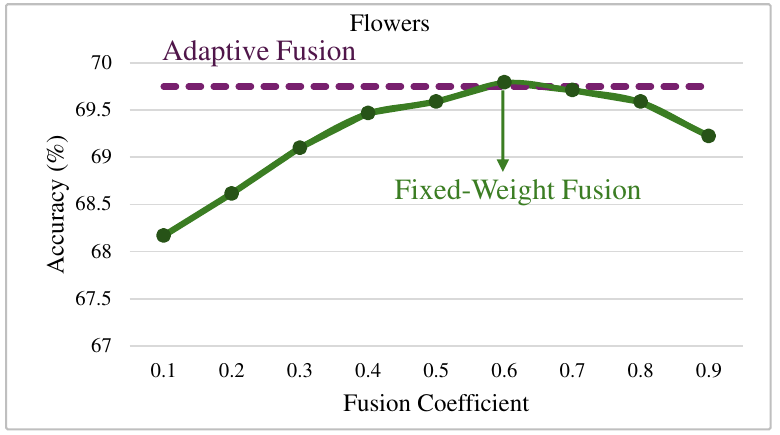}
            \vspace{0.1cm}
            \caption*{(b) Flowers}
        \end{minipage} \\

        \begin{minipage}{0.4\textwidth}
            \centering
            \includegraphics[width=\linewidth]{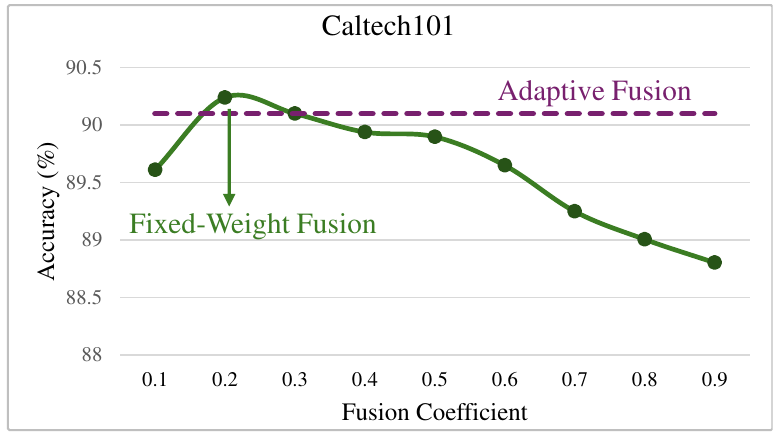}
            \vspace{0.1cm}
            \caption*{(c) Caltech}
        \end{minipage} &
        \begin{minipage}{0.4\textwidth}
            \centering
            \includegraphics[width=\linewidth]{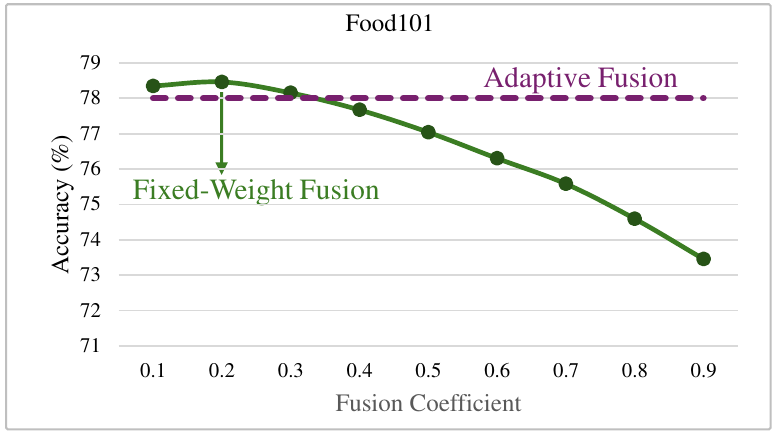}
            \vspace{0.1cm}
            \caption*{(d) Food101}
        \end{minipage} \\

        \begin{minipage}{0.4\textwidth}
            \centering
            \includegraphics[width=\linewidth]{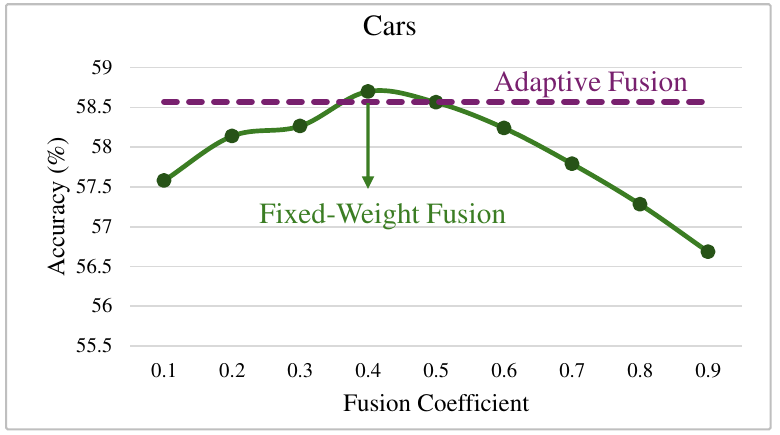}
            \vspace{0.1cm}
            \caption*{(e) Cars}
        \end{minipage} &
        \begin{minipage}{0.4\textwidth}
            \centering
            \includegraphics[width=\linewidth]{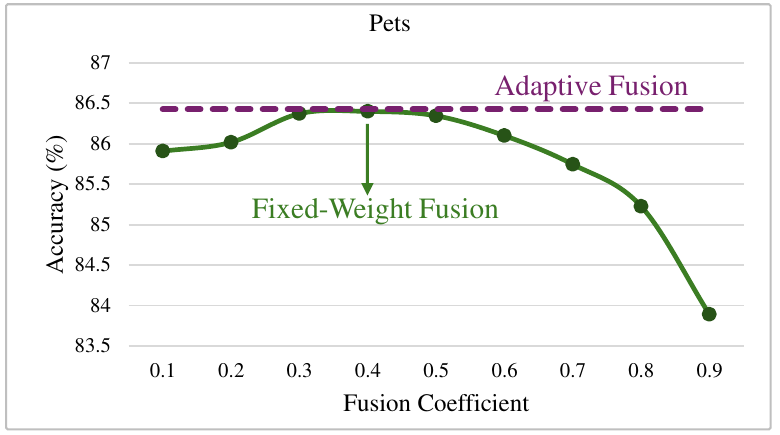}
            \vspace{0.1cm}
            \caption*{(f) Pets}
        \end{minipage} \\

        \begin{minipage}{0.4\textwidth}
            \centering
            \includegraphics[width=\linewidth]{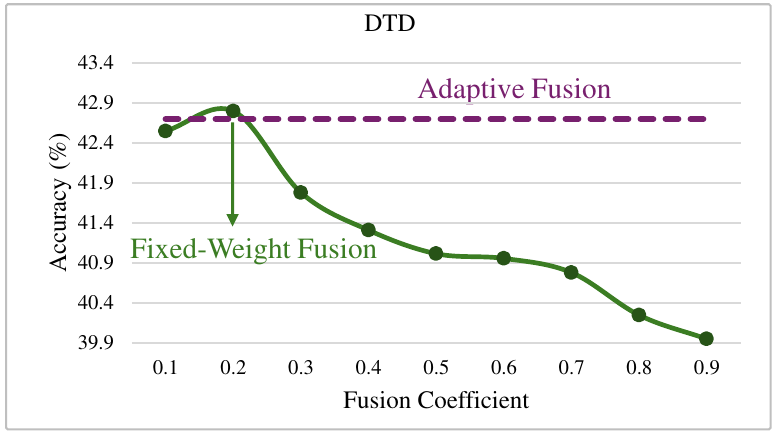}
            \vspace{0.1cm}
            \caption*{(g) DTD}
        \end{minipage} &
        \begin{minipage}{0.4\textwidth}
            \centering
            \includegraphics[width=\linewidth]{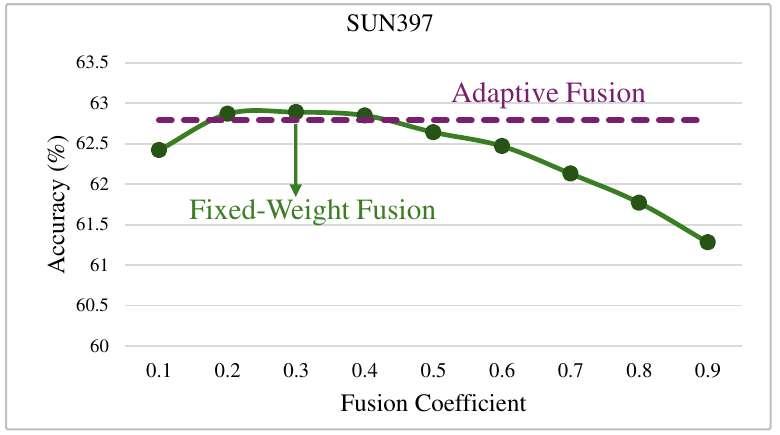}
            \vspace{0.1cm}
            \caption*{(h) SUN397}
        \end{minipage} \\

        \begin{minipage}{0.4\textwidth}
            \centering
            \includegraphics[width=\linewidth]{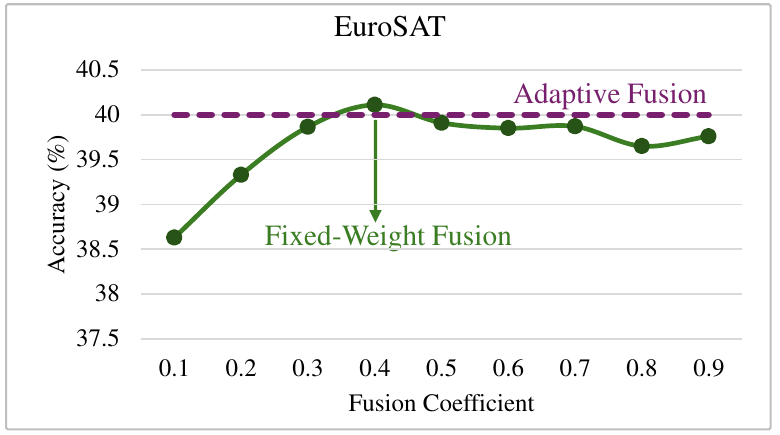}
            \vspace{0.1cm}
            \caption*{(i) EuroSAT}
        \end{minipage} &
        \begin{minipage}{0.4\textwidth}
            \centering
            \includegraphics[width=\linewidth]{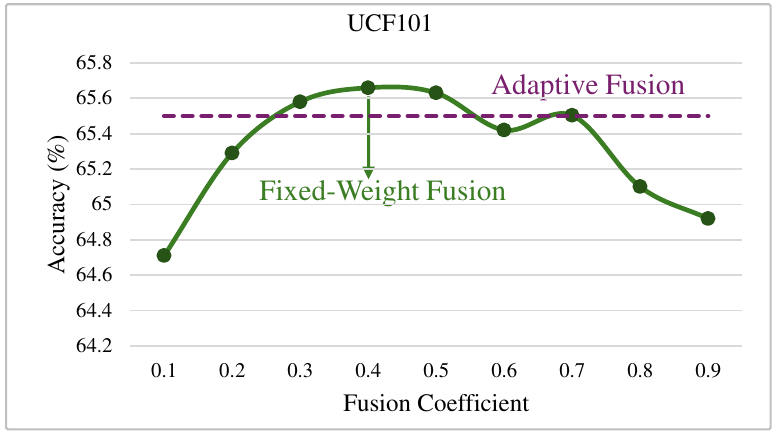}
            \vspace{0.1cm}
            \caption*{(j) UCF101}
        \end{minipage} \\
    \end{tabular}
\vspace{0.1cm}
    \caption{\small Performance comparison of the Adaptive Fusion module across different datasets. Each subfigure represents a specific dataset: (a) Aircraft \cite{aircraft}, (b) Flowers \cite{flower}, (c) Caltech \cite{caltech}, (d) Food101 \cite{food}, (e) Cars \cite{cars}, (f) Pets \cite{pets}, (g) DTD \cite{dtd}, (h) SUN397 \cite{sun397}, (i) EuroSAT \cite{eurosat}, and (j) UCF101 \cite{ucf101}.}
    \label{fig:adaptive-fusion-arranged}
\end{figure}

To evaluate the effectiveness of the Adaptive Fusion module, we conducted an ablation study in Section \ref{sec:ablation} using the ImageNet \cite{imagenet} validation dataset, comparing it with the Fixed-Weight Fusion strategy. To ensure generalizability, we extended the experiments to all cross-domain datasets using the ResNet-50 backbone. As shown in Fig. \ref{fig:adaptive-fusion-arranged}, the performance of Fixed-Weight Fusion depends heavily on the choice of fusion coefficients, with the optimal value varying significantly across datasets (e.g., Flower102 \cite{flower} achieves maximum accuracy at \(\beta = 0.6\), Food101 \cite{food} at \(\beta = 0.2\), and EuroSAT \cite{eurosat} at \(\beta = 0.4\)). In contrast, the Adaptive Fusion module dynamically adjusts weights based on the confidence of the logits, achieving accuracy consistently close to the optimal Fixed-Weight Fusion without requiring manual tuning. This adaptive approach eliminates the need for dataset-specific parameter adjustments, making it a robust and generalizable solution for combining logits across diverse downstream tasks.

\pagebreak
\newpage
\section{Robustness to Noisy Pseudo-Labels}
\label{sec:robustness}
This section evaluates the robustness of ETTA against noisy pseudo-labels, comparing it to TDA \cite{TDA} using the Flower dataset \cite{flower}. The analysis is conducted under varying levels of noise, denoted as \( \sigma \), to understand how both methods handle pseudo-label degradation.  

To assess robustness, ETTA and TDA are tested across different cache sizes and image noise levels. In a cache-based setting, increasing the cache size allows more prototypes to be stored, but it also introduces a higher likelihood of incorporating noisy prototypes. Similarly, higher noise levels negatively impact pseudo-labels, making them less reliable for adaptation. The experiment aims to determine how well each method can maintain stable performance in these challenging conditions.  

\begin{figure}[!t]
    \centering
    \includegraphics[width=0.9\linewidth]{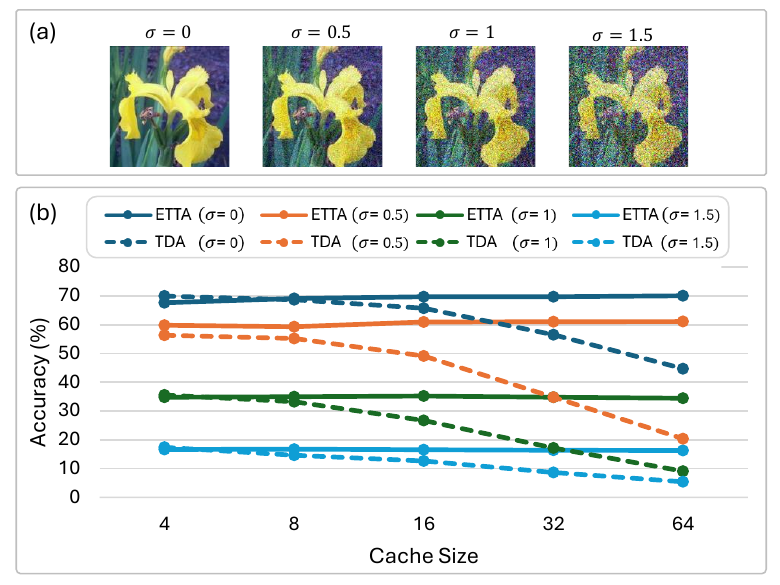}
    \vspace{0.2cm}
    \caption{\small Comparison of ETTA and TDA \cite{TDA} under varying levels of pseudo-label noise on the Flower dataset \cite{flower}. The x-axis represents different cache sizes, while the y-axis shows classification accuracy. As noise increases, TDA degrades significantly due to its reliance on stored prototypes, while ETTA remains stable, demonstrating the robustness of its Recursive Update module in mitigating the impact of noisy pseudo-labels.}
    \label{fig:cache-Compare-noisyl}
\end{figure}

The results, visualized in Fig.~\ref{fig:cache-Compare-noisyl}, show a direct comparison of ETTA and TDA under various noise levels. Additionally, Fig.~\ref{fig:cache-Compare-noisyl} examines how performance changes with increasing cache sizes. The findings reveal that TDA degrades significantly as the cache size increases, indicating that its reliance on stored prototypes makes it more vulnerable to pseudo-label noise. In contrast, ETTA remains stable across different cache sizes and noise levels. This highlights the effectiveness of its Recursive Update module, which helps mitigate the impact of noise by continuously refining the prototypes rather than passively accumulating them.  

\end{document}